\documentclass{article}

\usepackage{PRIMEarxiv}

\usepackage[utf8]{inputenc} 
\usepackage[T1]{fontenc}    
\usepackage{hyperref}       
\usepackage{url}            
\usepackage{booktabs}       
\usepackage{amsfonts}       
\usepackage{nicefrac}       
\usepackage{microtype}      
\usepackage{lipsum}
\usepackage{fancyhdr}       
\usepackage{graphicx}       
\graphicspath{{media/}}     

\usepackage{url}            
\usepackage{booktabs}       
\usepackage{amsfonts}       
\usepackage{nicefrac}       
\usepackage{microtype}      
\usepackage{xcolor}         

\usepackage{amssymb}
\usepackage{amsmath}
\usepackage{mathrsfs}
\usepackage{dutchcal}
\usepackage{bbm}
\usepackage{tikz}
\usetikzlibrary{positioning}
\usetikzlibrary{bayesnet}
\usepackage{graphicx}
\usepackage{subcaption}
\usepackage{caption}
\usepackage{wrapfig}

\title{Time-to-Pattern: Information-Theoretic Unsupervised Learning for Scalable Time Series Summarization}

\author{
  Alireza Ghods, Trong Nghia Hoang, Diane J. Cook \\
  Department of Computer Science \\
  Washington State University \\
  Pullman, WA 99164-1227, USA\\
  \texttt{\{alireza.ghods, trongnghia.hoang, djcook\}@wsu.edu} \\
}

\begin{document}

\maketitle

\begin{abstract}
Data summarization is the process of generating interpretable and representative subsets from a dataset. Existing time series summarization approaches often search for recurring subsequences using a set of manually devised similarity functions to summarize the data. However, such approaches are fraught with limitations stemming from an exhaustive search coupled with a heuristic definition of series similarity. Such approaches affect the diversity and comprehensiveness of the generated data summaries. To mitigate these limitations, we introduce an approach to time series summarization, called Time-to-Pattern (T2P), which aims to find a set of diverse patterns that together encode the most salient information, following the notion of minimum description length. T2P is implemented as a deep generative model that learns informative embeddings of the discrete time series on a latent space specifically designed to be interpretable. Our synthetic and real-world experiments reveal that T2P discovers informative patterns, even in noisy and complex settings. Furthermore, our results also showcase the improved performance of T2P over previous work in pattern diversity and processing scalability, which conclusively demonstrate the algorithm's effectiveness for time series summarization.
\end{abstract}

\section{Introduction}

The rapid proliferation of IoT sensors and online data collection mechanisms~\cite{byabazaire2020data} has led to remarkable growth in the amount and availability of vast, diverse time series data. Consequently, extracting meaningful patterns from complex time series data is increasingly crucial for effectively interpreting these resources~\cite{ahmed2019data}. Data summarization aims to produce a simplifying report that comprehensively describes the data~\cite{el2014interpretable,wen2018interactive,youngmann2022guided}. A helpful summary contains a set of descriptive patterns. For example, a pattern can be defined as either a direct subsequence of the time series or its latent embedding, which summarizes its most salient information and can be decoded back to the original space of the time series.


In this view, each pattern is expected to accurately reflect an important part of the original data (pattern {\em fidelity}) either as a direct subsequence of data or via its decoded representation. In addition, patterns must also be sufficiently diverse in content to represent all aspects of the time series, rather than just the most frequent ones (pattern {\em diversity}) ~\cite{wen2018interactive,youngmann2022guided,yu2009takes}. These are all important desiderata as time series summarization plays a critical role in numerous tasks, including data management~\cite{fan2000summary}, data interpretation~\cite{hooi2017b,Imani2020IntroducingTS}, and boosting the performance of machine learning algorithms~\cite{hooi2017b} with increased computational efficiency~\cite{plessen2020integrated}.

Despite the numerous approaches that have been attempted~\cite{hooi2017b,Imani2020IntroducingTS,Keogh2004ClusteringOT,Ye2009TimeSS,Zhu2016MatrixPI}, limitations still must be addressed in identifying informative subsequences. Owing to their exhaustive search, these methods grapple with scalability issues, and their dependence on a similarity function inevitably ushers in a degree of bias. Instead, we propose a new approach that is based on a neural network method that does not rely on exhaustive search and benefits from GPU and TPU hardware for better scalability. Furthermore, we tackle this problem by identifying patterns within the data that maximize information compression, following the principles of information theory. In particular, we adopt the Minimum Description Length (MDL) principle~\cite{wallace1968information} which asserts that an optimal summarizing pattern is the one that most greatly compresses the data. We hypothesize that we can formulate a learning algorithm to identify a diverse set of time series patterns that embody this principle.

To achieve this, we develop a framework called {\it time-to-pattern} (T2P), that models time series using a variational autoencoder (VAE) which is parameterized by a pair of neural encoder and decoder. Among the information-theoretic methods considered for data compression, VAEs have emerged as some of the most powerful for statistical modeling~\cite{seninge2021vega}. T2P's sparse latent space and limited-capacity decoder bolster learning efficiency and interpretability while capturing high-level abstract features. This approach learns disentangled representations of the data that compress the time series with fidelity and diversity. Because the encoder provides similarity scores between input sequences and learned patterns, the model's reasoning is more transparent.
Through comparative evaluation, we showcase T2P's improved data compressibility, pattern diversity, model interpretability, and algorithm scalability compared to state-of-the-art time series pattern mining. The main contributions of this work are:
\begin{itemize}
    \item {We introduce an information-theoretic approach to time series summarization, identifying patterns that offer compressibility, fidelity, and diversity while ensuring scalability.}
    \item {We design a learnable VAE with a Bernoulli prior, which does not rely on a fixed distance function to discover patterns but decomposes the latent encoding of time series into disentangled patterns, more effectively spanning the information spectrum.}
    \item {We conduct experiments that affirm T2P's superior performance over baseline methods for synthetic and real time series.}
\end{itemize}

\section{Background}

\paragraph{Minimum Description Length} T2P's approach draws from  information theory's minimum description length (MDL) principle~\cite{rissanen1978modeling}, which affirms that the best explanation of a theory, given limited observed data, is the one that maximally compresses the description length (DL) of the data. Thus, the optimal pattern $\mathcal{p} \in \mathbcal{P}$ to summarize time series $T$ is the one that minimizes the number of bits needed to describe $T|\mathcal{p}$ (the description length of $T$ where each pattern occurrence is replaced by a pointer to the pattern definition) plus the description of $\mathcal{p}$, or $\min_{\mathcal{p} \in \mathbcal{P}} DL(T|\mathcal{p}) + DL(\mathcal{p})$. For clarity, let's consider an example. Imagine a time series of length 1000 that contains two distinct patterns. Each pattern, let's call them pattern A and pattern B, has a length of 100. If pattern A and pattern B appear 5 times each in the time series, then $DL(T | \mathcal{P})$ is 10. This is because the time series can be described using 5 instances of pattern A and 5 instances of pattern B. The description length for the patterns, $DL(\mathcal{P})$, is the sum of the lengths of pattern A and pattern B, which equals 200. The pattern which minimizes this formula also effectively maximizes compression of the original time series $T$. We use the notion of minimum description length as an evaluation metric in our experiment (see Eq.~\eqref{eq:dl}).

\paragraph{Information-Theoretic Interpretation of VAEs} T2P harnesses the expressiveness of a Variational Autoencoder (VAE). VAEs were introduced as a deep unsupervised approach to learning latent data representations~\cite{Kingma2013AutoEncodingVB, JimenezRezende2014StochasticBA} by combining deep networks with Bayesian inference. A VAE consists of an encoder and a decoder. The encoder maps data $\mathbf{x}$ to a latent variable $\mathbf{z}$ by approximating the true posterior distribution $p(\mathbf{z}|\mathbf{x})$ with a variational distribution $q(\mathbf{z}|\mathbf{x};\boldsymbol{\varphi})$, where $\boldsymbol{\varphi}$ denotes the encoder's parameters. The decoder reconstructs the original data by modeling the likelihood $p(\mathbf{x}|\mathbf{z}; \boldsymbol{\theta})$, where $\boldsymbol{\theta}$ represents the decoder's parameters.
\begin{eqnarray}
\mathcal{L}(\boldsymbol{\theta}, \boldsymbol{\varphi}; \mathbf{x}) &=& \mathbb{E}_{q(\mathbf{z}|\mathbf{x};\boldsymbol{\varphi})}[\log p(\mathbf{x}|\mathbf{z}; \boldsymbol{\theta})] - \mathrm{KL}\Big(q(\mathbf{z}|\mathbf{x}; \boldsymbol{\varphi}) \parallel p(\mathbf{z})\Big) \label{eq:ELBO}
\end{eqnarray}
The VAE's objective is to minimize the Evidence Lower BOund (ELBO), a lower bound on the log-likelihood of the data. In Equation~\ref{eq:ELBO}, the first term represents the reconstruction loss, or {\it fidelity} of the data reconstructed from the latent variable. The second term denotes the Kullback-Leibler (KL) divergence between the approximate posterior $q(\mathbf{z}|\mathbf{x};\boldsymbol{\varphi})$ and the prior distribution $p(\mathbf{z})$, a regularization term that encourages the learned latent space to match the prior.

Many standard VAEs assume a multivariate Gaussian distribution for the prior $p(\mathbf{z})$. In contrast, T2P employs a Bernoulli-type distribution to support learning diverse patterns using a disentangled sparse latent space. In this approach, the sparse latent space serves as a switch, informing the decoder which learned pattern should be utilized for input reconstruction. By opting for the Bernoulli distribution, we promote sparsity in the latent space, aiding in the acquisition of {\it a diverse pattern set}. Sparsity serves as a form of regularization, compelling the model to depict data with fewer dimensions. In T2P, the decoder is structured as a function that takes the latent space as input to produce outputs. Such a sparse latent space ensures that the decoder learns a more varied set of kernels or features. Every active dimension in the latent space must account for a significant portion of the data's variability, potentially prompting the model to identify more unique features. Furthermore, a sparse latent space is often more interpretable. When only a few dimensions are active for a given input, it implies that those dimensions are crucial for that specific representation.

As a discrete distribution, the Bernoulli exhibits non-differentiability, necessitating the implementation of a differentiable reparametrization. We selected the BinConcrete distribution~\cite{Maddison2016TheCD} as a continuous relaxation of the Bernoulli distribution obtained through the Gumbel-Softmax trick that closely imitates Bernoulli~\cite{gumbel1954statistical,maddison2014sampling}, with a temperature parameter, $\boldsymbol{\lambda}$, that regulates the relaxation extent. BinConcrete models a continuous random variable with a probability density function that is differentiable with respect to its parameters.

The VAE can be interpreted through the lens of the {\em minimum description length} principle. To do this, we rearrange terms in the ELBO equation as shown in Equation~\ref{eq:itELBO}, where the first term probabilistically reflects the number of bits needed to construct data $\mathbf{x}$ using the model. This term is constant with respect to $\boldsymbol{\varphi}$ and $\boldsymbol{\theta}$. The KL divergence term quantifies the additional information that is required to represent samples from $q(\mathbf{z}|\mathbf{x}; \boldsymbol{\varphi})$ when using an optimal code for samples from $p(\mathbf{z}|\mathbf{x}; \boldsymbol{\theta}, \boldsymbol{\varphi})$. Minimizing this divergence corresponds to finding an efficient encoding for the latent variable $\mathbf{z}$ that maximally compresses the information content in $\mathbf{x}$ while enabling accurate reconstruction.
\begin{eqnarray}
    \mathcal{L}(\boldsymbol{\theta}, \boldsymbol{\varphi}; \mathbf{x}) &=& \log p(\mathbf{x}) \ -\ \mathrm{KL}\Big(q(\mathbf{z}|\mathbf{x}; \boldsymbol{\varphi}) \parallel p(\mathbf{z}|\mathbf{x}; \boldsymbol{\theta}, \boldsymbol{\varphi})\Big)
    \label{eq:itELBO}
\end{eqnarray}
The ELBO terms also relate to the concepts of rate and distortion~\cite{shannon1959coding}. \(\mathbb{E}_{q(\mathbf{z}|\mathbf{x};\boldsymbol{\varphi})}[\log p(\mathbf{x}|\mathbf{z}; \boldsymbol{\theta})]\) reflects the negative distortion, which measures the fidelity of the data reconstruction from the latent variable. \(\mathrm{KL}(q(\mathbf{z}|\mathbf{x}; \boldsymbol{\varphi}) \parallel p(\mathbf{z}))\), reflects the rate, or the amount of information that is needed to represent the latent variable \(\mathbf{z}\) given the input data \(\mathbf{x}\). In this context, VAE optimization can be viewed as finding an ideal trade-off between rate and distortion. If the rate is too high, the model may overfit the data, capturing data noise instead of learning a compact and meaningful representation. Conversely, if the rate is too low, the model may underfit the data, resulting in poor reconstruction quality. By optimizing the ELBO, VAEs learn to compress the time series into a lower-dimensional latent space while preserving the ability to generate high-quality reconstructions.

\section{Related Work}
\label{sec:related}

\paragraph{Time Series Pattern Discovery}
Search-based pattern discovery has emerged as a predominant focus of time series research. To mitigate the corresponding computational cost, researchers transform time series data into more compact representations~\cite{Foutsos1994FastSM,keogh2005exact,Keogh2004ClusteringOT,Lin2003ASR,Popivanov2002SimilaritySO}. 
However, these strategies experience information loss~\cite{song2020transitional}, sensitivity to hyperparameters~\cite{chen2023efficient} and data noise~\cite{rezvani2019new}, limited expressiveness, and the inability to handle nonlinear patterns~\cite{li2022new}.

Data summarization employs similar methods to discover similar sequence pairs. Termed motif discovery~\cite{Keogh2004ClusteringOT, Ye2009TimeSS, Zhu2016MatrixPI}, such techniques rely on exhaustive search. Thus, they aim to minimize the computational cost of computing sequence similarity, the core operation. Matrix Profiles (MPs)~\cite{Yeh2016MatrixPI} offer one such solution. Each entry in the MP denotes the smallest distance between a subsequence and its closest non-overlapping counterpart. A widely-used MP calculates the z-normalized Euclidean distance (ED) between all pairs of fixed-length subsequences~\cite{Yeh2016MatrixPI, lu2022matrix}. To broaden the spectrum of identifiable patterns, ED can be substituted with the more computationally expensive Dynamic Time Warping (DTW)~\cite{Alaee2021TimeSM}. Recently, the parameter-free Spikelet has been introduced to discover motifs that vary in scale by mapping continuous data to symbolic approximations and creating an MP based on this new representation and DTW~\cite{imamura2023parameter}.

\begin{figure}[h]
    \centering
    \begin{subfigure}{.36\textwidth}
        \centering
        \includegraphics[scale=0.18]{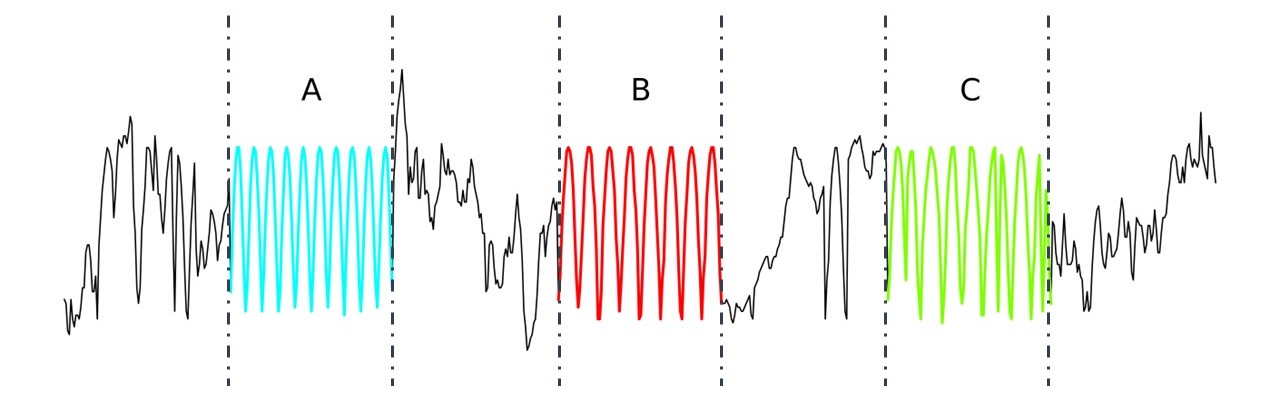}
        \caption{Synthetic time series.}
        \label{fig:EDcluster}
    \end{subfigure}
    \hspace{1em}
    \begin{subfigure}{.29\textwidth}
        \centering
        \includegraphics[height=3.5cm]{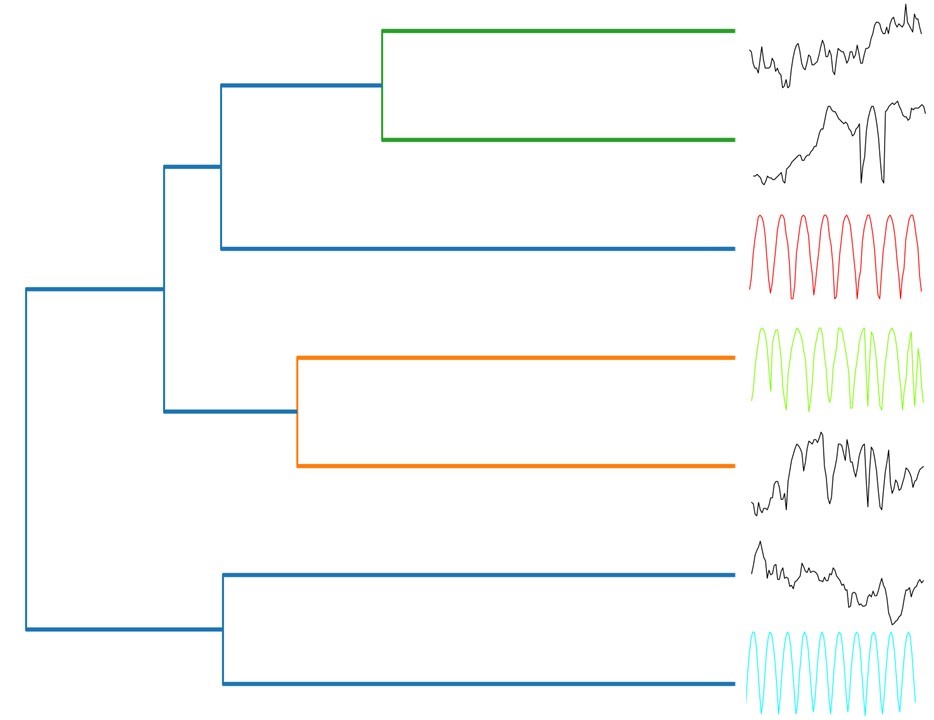}
        \caption{ED Cluster dendogram.}
        \label{fig:dendrogram}
    \end{subfigure}
    \vspace{0.1em}
    \begin{subfigure}{.29\textwidth}
        \centering
        \includegraphics[height=3.5cm]{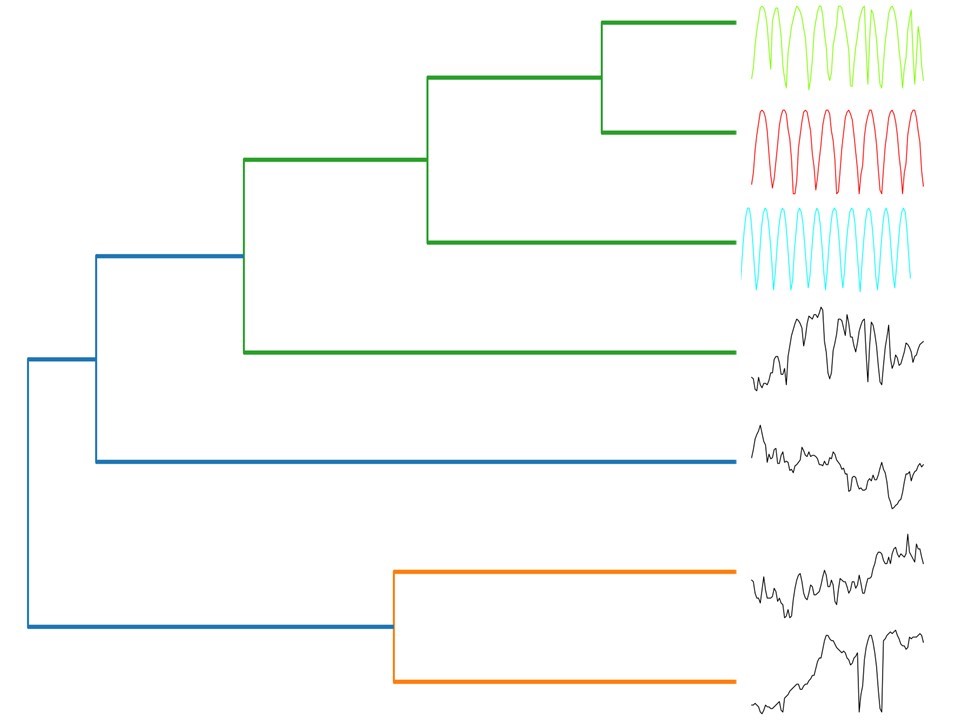}
        \caption{DTW cluster dendogram.}
        \label{fig:DTWdendrogram}
    \end{subfigure}
    \caption{A synthetic time series divided into seven equal-length regions and clustered using complete linkage hierarchical clustering.} 
    \label{fig:sim}
\end{figure}

However, these methods fall short of discovering patterns that most effectively and efficiently summarize the data. Consider a dataset that contains three periodic elements embedded into a random walk time series, as illustrated in Figure \ref{fig:EDcluster}. Pattern {\bf \color{cyan} A} contains ten periods, pattern {\bf \color{red}B} contains eight, and pattern {\bf \color{green} C's} ten periods are warped. We divided the dataset into seven equal-length regions and applied hierarchical clustering, as shown in Figure \ref{fig:dendrogram}. The three periodic elements are no more similar to each other under ED than they are to the random walks, as evidenced by the dendrogram. Although DTW accounts for warping, it surprisingly places pattern {\bf \color{green} C} closer to {\bf \color{red}B} than {\bf \color{cyan} A}. While DTW-driven search methods struggle with large and diverse time series, the deep-network T2P actually benefits from large datasets. Moreover, applying DTW to large time series may be impractical due to the prohibitive computational cost.\footnote{Our proposed method, T2P, does not exhibit the same limitations. The results of T2P applied to these data can be found in Online Appendix 1 at \url{https://github.com/alirezaghods/T2P-Time-to-Pattern/blob/main/Appendix.pdf}.}



An alternative data summarization approach is found in the more recent MP-Snippets~\cite{Imani2020IntroducingTS} algorithm. MP-Snippets employs local minima identification followed by the aggregation of analogous subsequences. This process ensures a holistic yet non-redundant encapsulation of the predominant patterns. While MP-Snippets is successful in finding representative patterns that summarize the data, the method still relies on a similarity function. These rule-based similarity-driven methods are prone to discovery bias and a tendency to overfit noise.\footnote{Additional explanations and experiments demonstrating the relationship between the choice of similarity function and discovered patterns are provided, together with T2P code and sample datasets, on the github site at \url{https://github.com/alirezaghods/T2P-Time-to-Pattern}.}  

Leveraging the power of deep networks to learn a greater variety of patterns, Noering et al.~\cite{noering2021pattern} proposed a convolutional autoencoder that discretizes fixed-size subsequences into bins, then replaces each bin by its average value to create a pixelated grayscale image. However, the simplified data representation leads to a proclivity for identifying only simplistic patterns. Furthermore, the method cannot process raw time series nor aid with interpreting the learned model.

\paragraph{Learning Disentangled and Sparse Representations} Prior research has investigated VAE enhancements that learn disentangled and sparse representations~\cite{Fortuin2018SOMVAEID,higgins2017beta,Kirschbaum2018LeMoNADeLM,makhzani2013k,van2017neural}. Disentangled representations support learning multiple patterns (pattern diversity) by forcing latent units to be sensitive only to generative factors. At the same time, sparsity enhances model interpretability and the ability to capture abstract features. K-sparse autoencoders~\cite{makhzani2013k} enforce sparsity by allowing only top-k neurons to contribute to reconstruction. Similarly, $\beta$-VAE~\cite{higgins2017beta} learns disentangled patterns by balancing reconstruction loss and KL-divergence. Similar to T2P, VQ-VAE~\cite{van2017neural} and LeMoNADe~\cite{Kirschbaum2018LeMoNADeLM} learn disentangled sparse representations that compress input data. In VQ-VAE, both posterior and prior distributions are categorical. LeMoNADe, on the other hand, uses a discrete stochastic node to identify recurring video patterns. SOM-VAE passes the encoder output through a discretization bottleneck, then maps the result to the nearest element within the embedding to cluster videos~\cite{Fortuin2018SOMVAEID}.  

Like these prior methods, T2P aims to discover disentangled and sparse data representations. T2P improves upon prior methods by learning compressive, disentagled patterns with a VAE, the first reported method of this type. To encode temporal relationships, T2P takes advantage of a continuous Bernoulli distribution as the prior. Furthermore, T2P offers a unique ability to provide an interpretable latent space without additional network structure.

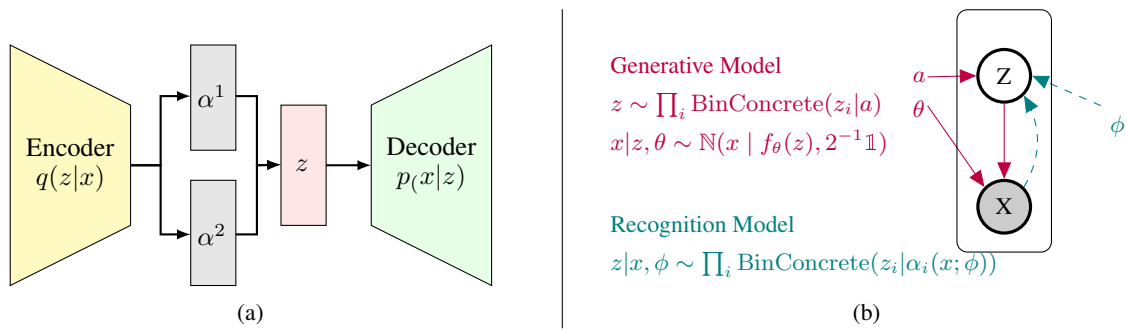
\begin{figure}[ht]
    \centering
        \begin{minipage}[b]{0.49\textwidth}
            \centering
            \begin{tikzpicture}[scale=0.80]
            \fill[yellow!30, draw=black, label=enc] (3,2) -- (3,-2) -- (5,-1) -- (5,1) -- cycle;
            \node[align=center] at (4,0) {Encoder\\$q(z|x)$};
            \fill[black!10, draw=black, label=a1] (6,0.25) rectangle (6.75,2);
            \node[align=center] at (6.35,1.15) {$\alpha^1$}; 
            \fill[black!10, draw=black, label=a2] (6,-0.25) rectangle (6.75,-2);
            \node[align=center] at (6.35,-1.15) {$\alpha^2$}; 
            \fill[red!10, draw=black, label=a2] (7.5,1) rectangle (8.25,-1);
            \node[align=center] at (7.85,0) {$z$}; 
            \fill[green!10, draw=black, label=enc] (9,1) -- (9,-1) -- (11,-2) -- (11,2) -- cycle;
            \node[align=center] at (10,0) {Decoder\\$p_(x|z)$};
            \draw[-latex, thick] (5,0) -- ++(0.5,0) -- ++(0,1.15) -- (6,1.15);
            \draw[-latex, thick] (5,0) -- ++(0.5,0) -- ++(0,-1.15) -- (6,-1.15);
            \draw[-latex, thick] (6.75,1.15) -- ++(0.35,0) -- ++(0,-1.15) -- (7.5,0.01);
            \draw[-latex, thick] (6.75,-1.15) -- ++(0.35,0) -- ++(0,+1.15) -- (7.5,0.01);
            \draw[-latex, thick] (8.25,0) -- (9,0);
        \end{tikzpicture}
        \subcaption{}
        \label{fig:framework}
    \end{minipage}
    \vrule
    \begin{minipage}[b]{0.49\textwidth}
        \centering
        \begin{tikzpicture}[x=6.7cm,y=2.4cm,
            roundnode/.style={circle, draw=black, fill=black!20, very thick, minimum size=7mm},
            roundnodeW/.style={circle, draw=black, very thick, minimum size=7mm},scale=0.80,font=\small,
            ] 
            \node[roundnode] (X)  {X};
            \node[roundnodeW] (Z) [above=of X] {Z};  
            \draw[purple,->] (Z.south) -- (X.north);
            \draw[teal, dashed, ->] (X.north east) to [bend right = 30] (Z.south east);          
            \draw [teal] (0.24,0.7) node[anchor=north west] {$\phi$};
            \draw[teal, dashed, ->] (.24,.7) to (Z.east);           
            \draw [purple] (-.25,.8) node[anchor=north west] {$\theta$};
            \draw [purple] (-.25,1.0) node[anchor=north west] {$a$};
            \draw[purple,->] (-.19,.9) to (Z.west);
            \draw[purple,->] (-.19,.7) to (X.north west);            
            \plate [inner sep=.25cm,xshift=0.0cm,yshift=.25cm] {plate1} {(X)(Z)} {};
            \draw [teal] (-1,0) node[anchor=north west] {Recognition Model};
            \draw [teal] (-1,-.25) node[anchor=north west] {$z|x,\phi \sim \prod_{i}\mathrm{BinConcrete}(z_i|\alpha_i(x;\phi))$};
            \draw [purple] (-1,1.1) node[anchor=north west] {Generative Model};
            \draw [purple] (-1,.85) node[anchor=north west] {$z\sim\prod_{i}\mathrm{BinConcrete}(z_i|a)$};
            \draw [purple] (-1,.60) node[anchor=north west] {$x|z,\theta\sim\mathbb{N}(x\mid f_{\theta}(z),2^{-1}\mathbbm{1})$};
            \end{tikzpicture}
        \subcaption{}
        \label{fig:subplot_b}
    \end{minipage}
    \caption{\textbf{(a)} T2P architecture. \textbf{(b)} Plate diagram of generative and recognition models. Solid lines denote the generative model $p_{\theta}(x|z)$ and dashed lines denote the variational approximation $q_{\varphi}(z|x)$.}
    \label{fig:model}
\end{figure}

\section{Time to Pattern}
\label{sec:method}

We formalize the notion of an information-theoretic pattern and propose a scalable solution to discover them via an augmented VAE, which (1) optimizes a probabilistic embedding of the time series with minimal information loss and (2) extracts a diverse set of informative patterns from the latent embedding. The VAE is configured so the latent variable $z$ contain the pattern activations, and the decoder indicates pattern firing within the cells.

\paragraph{Information-Theoretic Pattern}
\label{sec:statement}
Let $\mathbf{x} \triangleq (x_1, x_2, \ldots, x_n)$ denote a time series and $\mathbcal{P}_\tau = \{\mathbf{x}^\dagger_1, \mathbf{x}^\dagger_2, \ldots, \mathbf{x}^\dagger_m\}$ represent a set of sequences of length $\tau < n$. $\mathbcal{P}_\tau$ is a set of {\em informative patterns} if $\mathbcal{P}_\tau$ is a sufficient statistic to infer any predictable (but unobservable) statistics $c(\mathbf{x})$ describing $\mathbf{x}$ (pattern {\it fidelity}). That is, $\mathbcal{P}_\tau$ satisfies $\mathbcal{P}(c(\mathbf{x})\mid \mathbcal{P}_\tau, \mathbf{x}) = \mathbcal{P}(c(\mathbf{x}) \mid \mathbcal{P}_\tau)$ or $c(\mathbf{x}) \perp \mathbf{x} \mid \mathbcal{P}_\tau$. Furthermore, for the principle of pattern {\it diversity} to hold, for all $i, j \in {1, 2, \ldots, k}$ and $i \neq j$, the constraint must hold that $\mathbf{x}^\dagger_i \neq \mathbf{x}^\dagger_j$.

\paragraph{T2P Framework}
The T2P architecture, depicted in Figure~\ref{fig:framework}, includes an encoder with a sequence of 1D convolutions capped off with a 3D convolution and a decoder featuring a 3D convolution layer. Details of the encoder/decoder structures are found in Appendix~\ref{appendix:t2p}. Utilizing the reparameterization trick, the encoder network maps the input to $\alpha^1$ and $\alpha^2$, both in the interval $(0, \infty)$, to derive the latent random variable $z \in {0,1}$. The value of $z$ is sampled from the prior distribution $p_{a}(z)$ for each subsequence $\mathbf{x}$, determining the activation or deactivation of pattern $\mathbcal{P}$. The number of nodes in the latent space and kernels in the decoder correspond to the user-defined number of desired patterns, $k$. The length of the discovered patterns is influenced by modifying the window size, $m$. Before training the model, the time series is segmented into non-overlapping subsequences of length $m$.

A crucial aspect of this framework is the single decoder layer with convolution. By constraining the decoder's capacity, T2P forces the layer to develop a set of kernels. Since the latent space is sparse, the model learns a collection of disentangled patterns from the data through these kernels. The primary objective of the decoder is to minimize the loss function by generating output that closely resembles the input. Because the decoder has limited capacity, it learns the most prevalent patterns that reduce the loss. This approach facilitates the extraction of highly representative patterns from the data, thereby contributing to the model's ability to learn patterns that closely resemble the input data.

\paragraph{Reparameterization Trick and Objective Function}
Let \( p_a(z) \) be a prior distribution over \( z \), where its parameter \( a \) represents a location within the prior distribution. The reparameterization trick samples \( z \) from the BinConcrete distribution as follows:

\begin{equation}
\hspace{-10mm}z \ \ =\ \ \text{softmax} \left( \frac{Y}{1+Y} \cdot\alpha^{1} \right)  \quad\text{with} \quad Y \ \ =\ \ \left(\frac{\alpha \cdot U}{1 - U + \epsilon}\right)^{1/\lambda_{1}}, \quad \text{and} \quad \alpha = \alpha^1 \oslash \alpha^2 
\label{eq:z} 
\end{equation}

where  \(\lambda_1\) is a temperature value, \( U \sim \text{Uniform}(0, 1) \), and \(\alpha\) results from element-wise division between \(\alpha^1\) and \(\alpha^2\). We incorporate Softmax and multiplication by \(\alpha^{1}\) to improve the result by approaching an almost-binary latent space in the T2P model, which sums to one. The sparse representation enhances interpretability by associating each latent feature with specific patterns, normalized to provide a probabilistic distribution of the patterns' contributions to the representation. 

Disentanglement is encouraged by setting the prior distribution location parameter $a$ close to one. We utilize a value between 0.8 and 0.9, encouraging the model to learn a sparser latent space. This enhanced model is more robust to variations in the input data, as the disentangled representation can better adapt to changes in individual patterns. Performance is improved due to the focus on essential features, promoting better generalization and computational efficiency. Since the model learns a probabilistic distribution over patterns, it can adapt to changes in the input data. Following the model parameterization in Fig.~\ref{fig:model}b, Eq. (\ref{eq:ELBO}) can now be re-written as\footnote{More details on the VAE are included in Online Appendix 2 at \url{https://github.com/alirezaghods/T2P-Time-to-Pattern/blob/main/Appendix.pdf}}

\begin{equation}
\text{MSE}(\mathbf{x}, \hat{\mathbf{x}}) \ \ +\ \  \mathrm{KL} \Big(g_{\alpha, \lambda_1}(Y|\mathbf{x}) \ \|\  f_{a, \lambda_{2}}(Y)\Big) \ ,
\label{eq:mse}
\end{equation}

where \(g_{\alpha, \lambda_1}(Y|\mathbf{x})\) represents the encoder network function which transforms the input \(\mathbf{x}\) into a latent variable \(Y\) with parameters \(\alpha\) and \(\lambda_1\), \(f_{a, \lambda_{2}}(Y)\) denotes the decoder network function, a generative model that recreates the original input from the latent variable \(Y\) with parameters \(a\) and \(\lambda_{2}\), $\text{MSE}(\mathbf{x}, \hat{\mathbf{x}})$ represents the mean squared error between input $\mathbf{x}$ and decoder output $\hat{\mathbf{x}}$, $g_{\alpha, \lambda_1}(Y|\mathbf{x})$ is the reparameterized Binary Concrete relaxation of the variational posterior $q_{\phi}(z|\mathbf{x})$ with temperature $\lambda_1$, and $f_{a, \lambda_{2}}(Y)$ is the density of a logistic random variable sampled via Eq.~\eqref{eq:z} with temperature $\lambda_2$ and the prior distribution location $a$. The KL-Divergence between prior and posterior distribution is

\begin{equation}
\mathrm{KL}(p(z) \| q(z|x)) = \int p(z) \log \frac{p(z)}{q(z \| x)} \,dx.
\end{equation}

To compute the KL divergence, we apply Monte Carlo Sampling. The probability density function for the BinConcrete distribution is given by

\begin{equation}
    \frac{\lambda \alpha x^{-(\lambda+1)}  (1 - x)^{-(\lambda+1)}}{(\alpha x^{-\lambda} + (1 - x)^{-\lambda})^2},
\end{equation}

where $\alpha$ represents the location of the distribution, $\lambda$ is the temperature parameter and variable $x$ ranges from 0 to 1. To compute the KL divergence, we can resort to Monte Carlo sampling, an effective numerical method that relies on random sampling to estimate mathematical expressions. Specifically, we draw $S$ random samples $z_i$ from the prior distribution $p(z)$, which is assumed to follow a uniform distribution.  For each sample $z_i$, we evaluate the log probability density under distributions $p(z)$ and $q(z|x)$:

\begin{align}
    \log p(z_i) & = \log f(z_i; a, \lambda_1) \\
    \log q(z_i|x) & = \log f(z_i; \alpha, \lambda_2)
\end{align}

We then compute the mean of the log probability density ratio for the samples:

\begin{equation}
    \mathrm{KL}(p(z) \| q(z|x)) \approx \frac{1}{S} \sum_{i=1}^{S} \left( \log p(z_i) - \log q(z_i|x) \right)
    \label{eq:kl}
\end{equation}

Combining Eq.~\eqref{eq:mse} and~\eqref{eq:kl}, T2P's objective is to minimize the value represented by Equation~\ref{eq:t2p}.

\begin{equation}
\text{MSE}(\mathbf{x},\mathbf{\hat{x}}) \ \ -\ \   \mathbb{E}_{Y \sim g_{\alpha, \lambda_{1}}(y|\mathbf{x})} \Bigg[ \log \frac{f_{a.\lambda_2 (Y)}}{g_{\alpha, \lambda_{1}}(Y|\mathbf{x})} \Bigg] \
\label{eq:t2p}
\end{equation}

Previous research has proposed modulating the effect of the KL-divergence by regulating the temperature value~\cite{higgins2017beta,Kirschbaum2018LeMoNADeLM}. However, this hyperparameter did not yield improvements in our experiments. The primary objective of T2P is to minimize the mean squared error between the input and decoder output while maximizing the expected value of the log-likelihood ratio of the two distributions, ultimately leading to pattern fidelity and diversity.\footnote{Details regarding the T2P architecture, training procedures, and parameter tuning can be found in Appendix A.}

\begin{figure}[ht]
    \centering
    \begin{subfigure}[t]{\textwidth}
        \centering
        \begin{tikzpicture}
        \node[anchor=south west, inner sep=0] (input) at (-1,0) {\includegraphics[width=6cm, height=2cm]{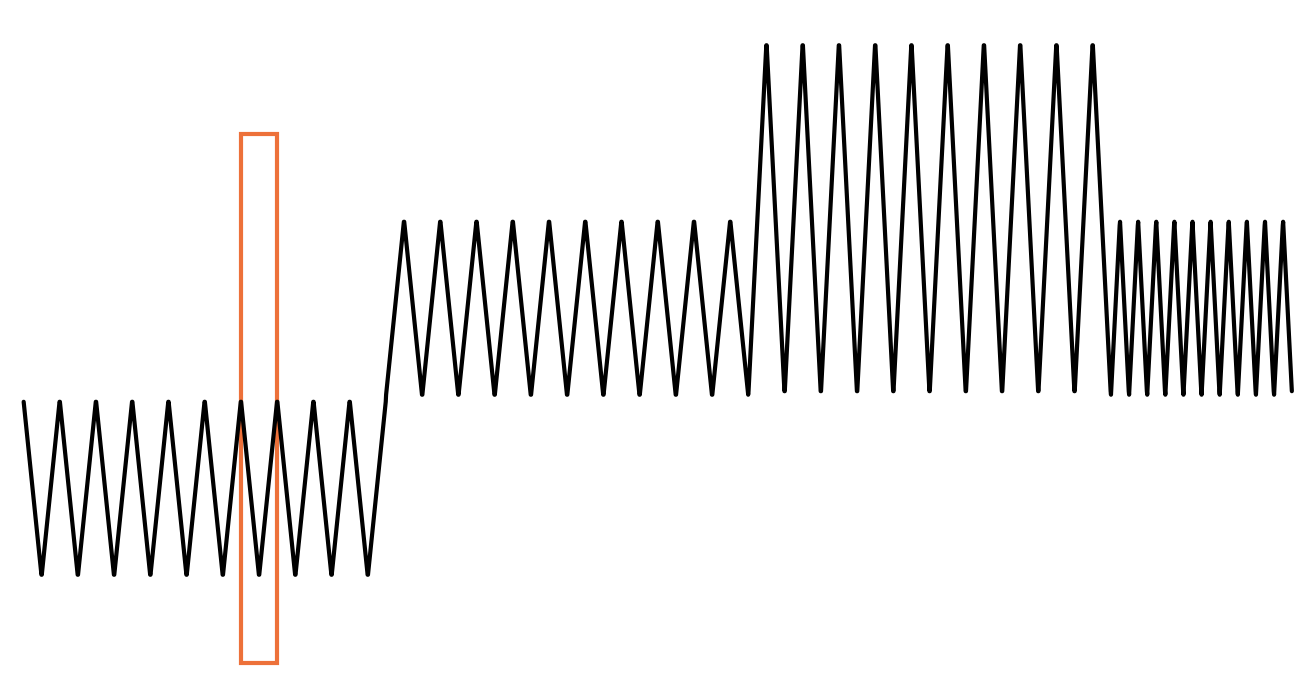}};
        \node[anchor=north] at (input.south) {Time series $T$};
        \node[anchor=south west, inner sep=0] (sub) at (6,0) {\includegraphics[width=0.9cm]{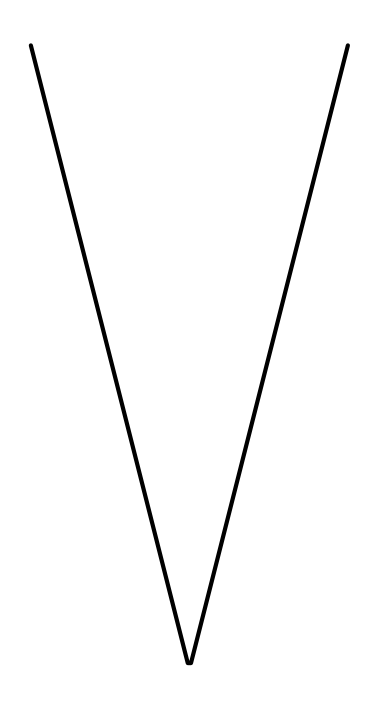}};
        \node[anchor=north] at (sub.south) {Subsequence $x$};
        \node[anchor=south west, inner sep=0] (a1) at (8,0) {\includegraphics[width=3.5cm]{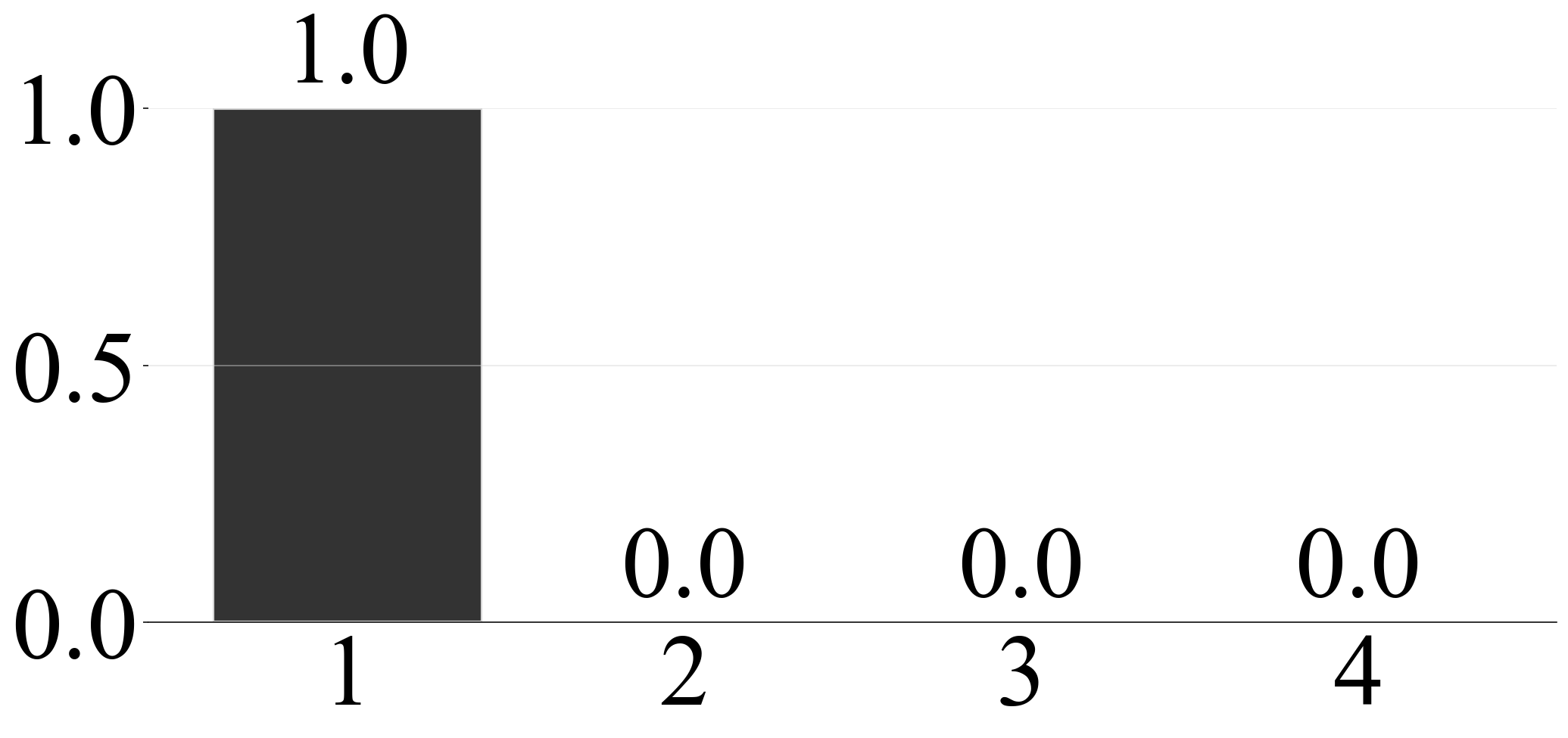}};
        \node[anchor=north] at (a1.south) {T2P latent space $z$};
        \draw[-latex, thick, orange] (.3,0.6) -- (6.,0.6);
        
        \end{tikzpicture}
        \caption{}
    \end{subfigure}
     \begin{subfigure}[b]{\textwidth}
        \centering
        \begin{tikzpicture}
        \node[anchor=south west, inner sep=0] (pattern) at (0,0){\includegraphics[width=3cm, height=2cm]{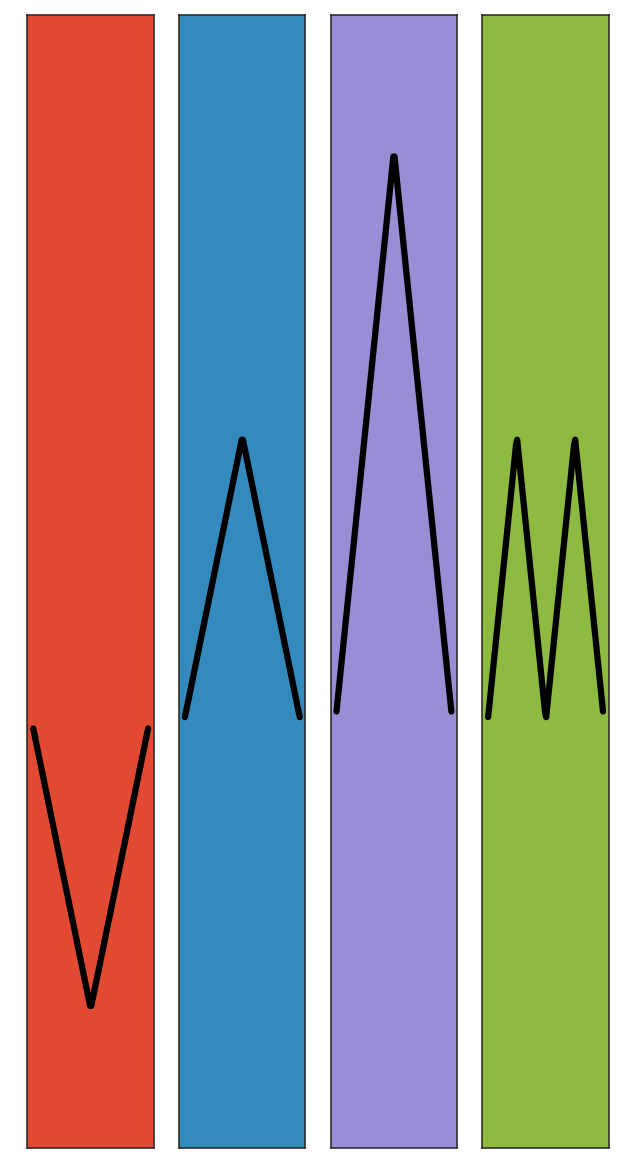}};
        \node[anchor=north] at (pattern.south) {Learned patterns};
        \node[anchor=south west, inner sep=0] (ana) at (7,0){\includegraphics[width=6cm,height=2cm]{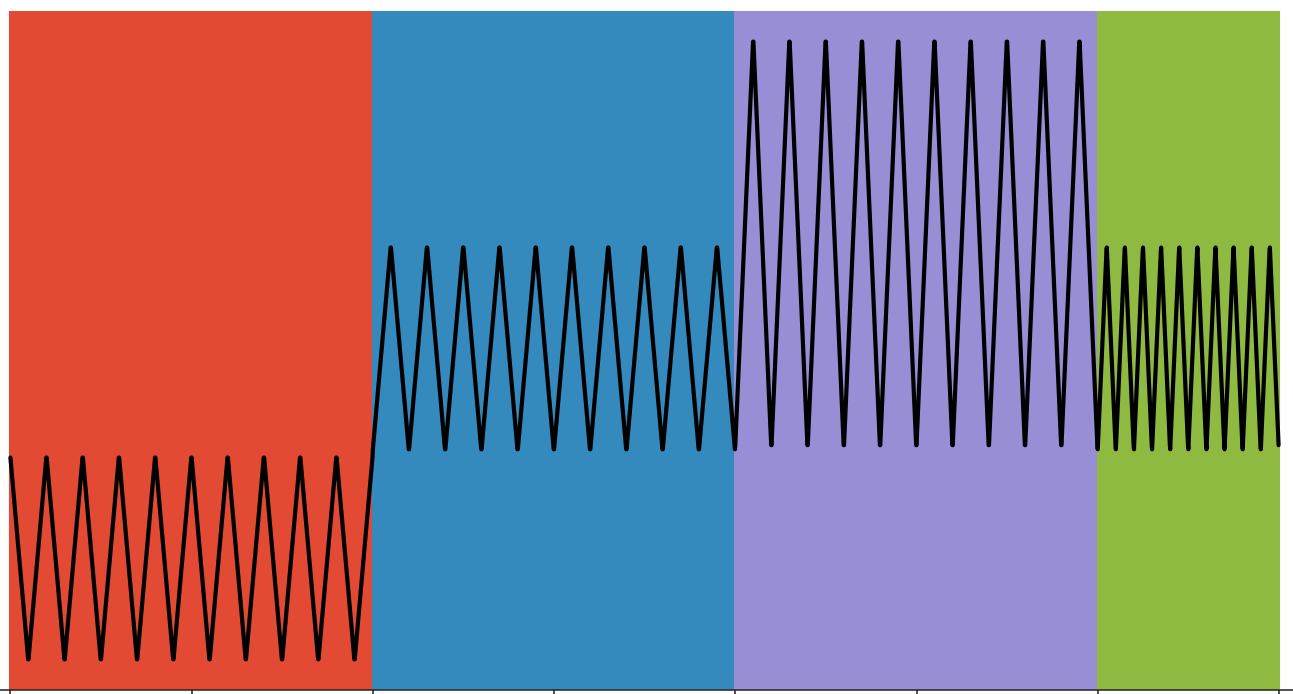}};
        \node[anchor=north] at (ana.south) {Identified pattern occurrences in $T$};
        \end{tikzpicture}
        \caption{}
        \label{fig:subplotB}
    \end{subfigure}
    \caption{When applying a T2P model to time series $T$, we can inspect the latent space, $z$, to ascertain which learned pattern aligns with each subsequence. (a) Subsequence $x$ aligns with the first element in vector $z$, characterized by the highest score. A score of 1 signifies an exact match between $x$ and the first pattern. (b) T2P's discovered patterns are depicted (left), and the relationships between $T$'s subsequences and the corresponding pattern are color mapped (right).}
    \label{fig:mechanism}
\end{figure}
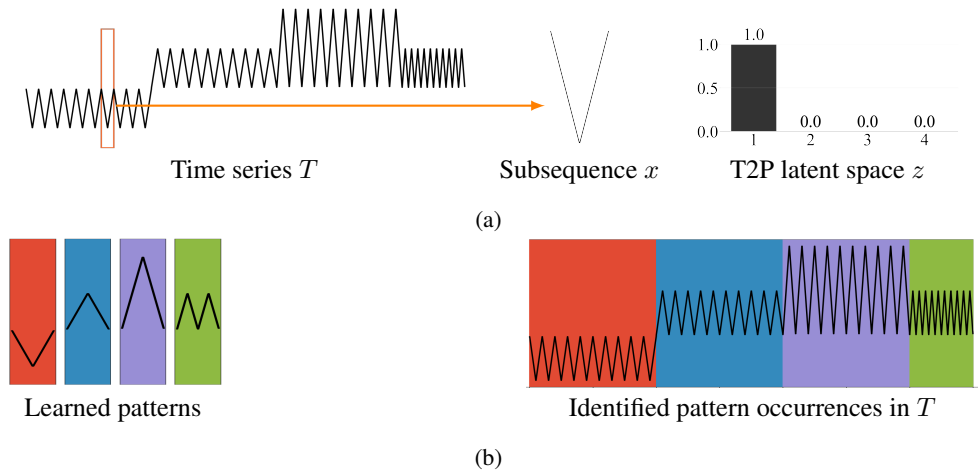

\paragraph{Interpreting T2P through Pattern Visualization} T2P employs a single decoder layer responsible for reconstructing the time series subsequences, $x_i$. Because we impose a limit on the number of convolution kernels, T2P must efficiently utilize these kernels for accurate data reconstruction. Consequently, T2P focuses on acquiring the most frequently occurring subsequences to minimize loss. Figure \ref{fig:mechanism} provides a graphical representation of these kernels for a trained model.  
Given a time series subsequence $x$ and the maximally-activated node in latent space $z_j$, the corresponding pattern, $\mathbcal{P}_j$, is selected to reconstruct the time series. The rationale is predicated on the fact that the latent space $z$ dictates the selection of the pattern employed for reconstruction, as the model computes the dot product of the latent space and deconvolution layer, $\sum_{i=1}^{k} z_i \cdot \mathbcal{P}_i$. As demonstrated in the ``activation map'' column in Figure \ref{fig:mechanism}.a (right), the first node in the latent space exhibits the largest activation, indicating that time series $x$ will be most accurately reconstructed from the first deconvolution kernel. In this example, T2P identifies a strong similarity between the input and the learned pattern, with a score of one. As shown in Figure~\ref{fig:mechanism}, the colors of the subseries indicate the corresponding identified pattern. The intensity of the color corresponds to the $z$ value for that pattern (in this case, $\forall x_i, \quad \max(z) = 1
$). This results in a similarity score map that elucidates the contribution of each pattern to the time series subsequences, boosting interpretability of T2P's summarization process.

\section{Experiments}

We assess the performance of T2P through a series of comparative experiments. There are few choices for baselines, because many of the methods described in Section~\ref{sec:related} do not address the same problem as T2P. We select MP-Snippets~\cite{Imani2020IntroducingTS} as a baseline in these experiments. Among summarization methods, MP-snippets emerges as a method that has demonstrated scalability and representation diversity. We select MP-Snippets because it shares a common goal of finding top-k patterns and does not lose fidelity by pre-compressing the data. MP-Snippets calculates the similarity between all subsequences using ED to identify the subsequences that best represent the data.  All experiments were run on an NVIDIA GeForce RTX 2070 GPU, an Intel Core i7-8700K CPU, and 32 GiB of memory, running Ubuntu 22.04.

We chose six datasets for these experiments. The first two datasets are constructed synthetically to evaluate algorithm performance for controlled conditions. The remaining three contain real-world data and vary in complexity. Specifically, the vital sign dataset has been utilized in previous time series analyses and thus offers a basis for comparison~\cite{Imani2020IntroducingTS}. The remaining four datasets represent real-world time series with increasing complexity. These represent audio signals~\cite{becker2018interpreting}, ECG readings~\cite{olszewski2001generalized}, and time series representations of airplane designs~\cite{thakoor2005hidden}.\footnote{Figures that visualize the results for each experiment are in Online Appendix 4 at \url{https://github.com/alirezaghods/T2P-Time-to-Pattern/blob/main/Appendix.pdf}.}

\subsection{Evaluation Metrics}

Measuring performance in unsupervised learning problems is inherently challenging. Consistent with the information-theoretic claim that patterns which maximally compress the data yield the best data summary, we first evaluate the compression performance by quantifying the compression of time series $T$ using pattern set $\mathbcal{P}$:

\begin{equation}
\frac{DL(T)}{DL(\mathbcal{P})+DL(T|\mathbcal{P})+DL(penalty)} \ ,
\label{eq:dl}
\end{equation}

where $DL(\mathbcal{P})$, $DL(T|\mathbcal{P})$, and $DL(penalty)$ denote the combined pattern lengths, the time series length post-pattern replacement, and the length of incorrectly-associated subsequences, respectively. Higher compression values indicate stronger performance.

To showcase the capabilities of our method, we further assess our results based on data that contain expert-identified patterns. 
In these cases, we define true positive (TP) as the fraction of data subsequences that are correctly paired with their corresponding patterns. False positive (FP) indicates the fraction of subsequences that are incorrectly paired, while false negative (FN) denotes the fraction of patterns that are present in the data but the algorithm failed to identify. We conduct each experiment five times and report the results as the mean $\pm$ standard deviation for identification precise and recall. We also empirically assess scalability. We illustrate in more detail hyperparameter effects, including  the number of patterns and pattern length, in Appendix C and the Online Appendix.

\subsection{Synthetic Data}
\label{section:synthetic}

{\em Embedding patterns with noise.} Time series summarization methods should effectively handle noise in the time series data originating from measurement error, system disturbances, or external influences, while retaining the time series' fundamental information. Details of how noise impacts T2P and other time series summarization methods are included in Appendix~\ref{app:dealingNoise}.

To evaluate T2P and MP-Snippets in terms of robustness to noise, we created four synthetic time series patterns, each of length 100 and characterized by different amplitudes and frequencies (see Figure~\ref{fig:subplotB}). Each time series is created by concatenating the four patterns. Next, we introduced noise into these patterns using a Gaussian noise model with noise levels ranging from 0\% (no noise) to 100\%. At each level, the corresponding amount of Gaussian noise was applied to the original time series.

As Table \ref{tab:comparison} shows, T2P consistently outperforms MP-Snippets across all metrics. For all time series, T2P patterns provide the greatest compression. Based on the known embedded patterns, T2P also provides the greatest accuracy in identifying pattern occurrences, as exemplified by the precision and recall values. As the noise level increases, MP-Snippets patterns bear less similarity to the embedded patterns, while T2P consistently preserves the embedded pattern's shape.

\begin{figure}[h]
    \centering
    \begin{subfigure}{0.45\textwidth}
        \centering
        \includegraphics[width=\textwidth]{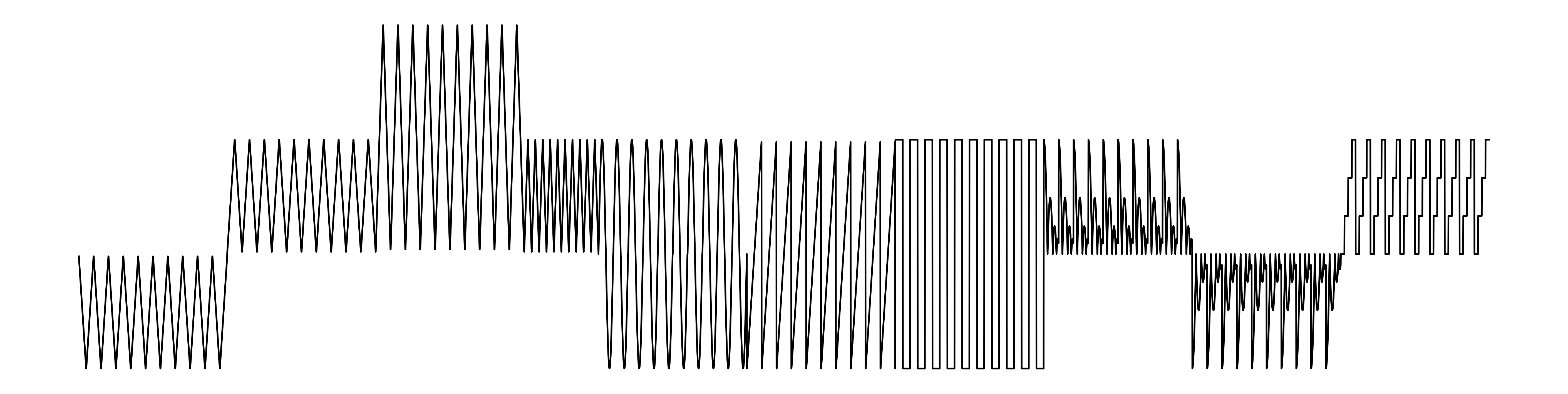}
        \caption{Synthetic data with ten patterns.}
        \label{fig:ten_a}
    \end{subfigure}
    \hspace{.1em} 
    \begin{subfigure}{0.45\textwidth}
        \centering
        \includegraphics[width=\textwidth]{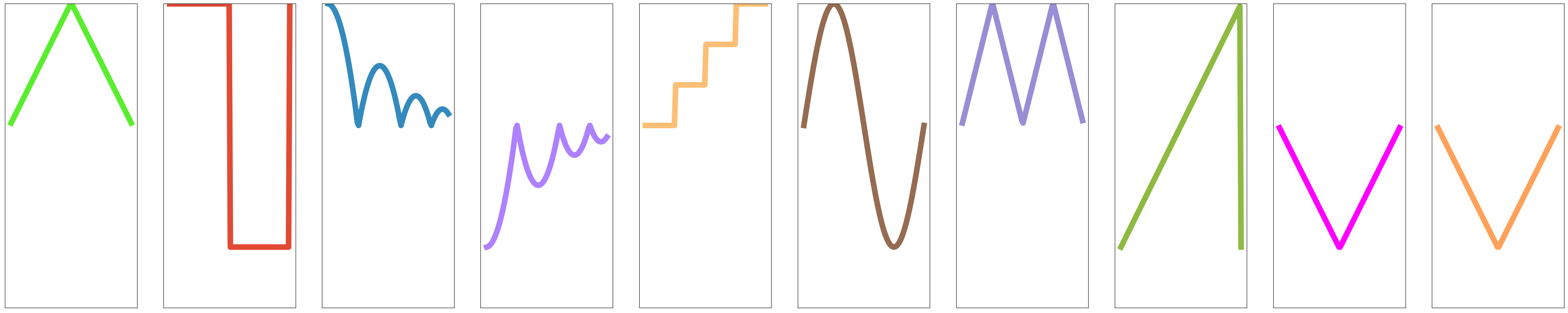}
        \caption{MP-Snippets top ten patterns.}
        \label{fig:ten_b}
    \end{subfigure}%
    \vspace{.1em} 
    \begin{subfigure}{0.45\textwidth}
        \centering
        \includegraphics[width=\textwidth]{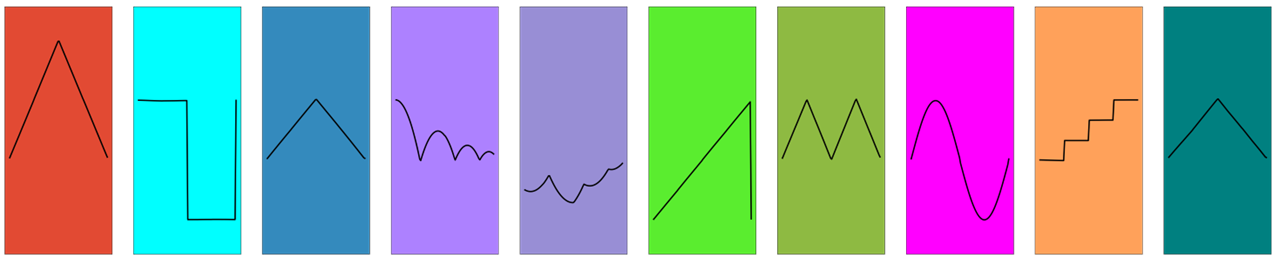}
        \caption{T2P ten patterns.}
        \label{fig:ten_c}
        \includegraphics[width=\textwidth]{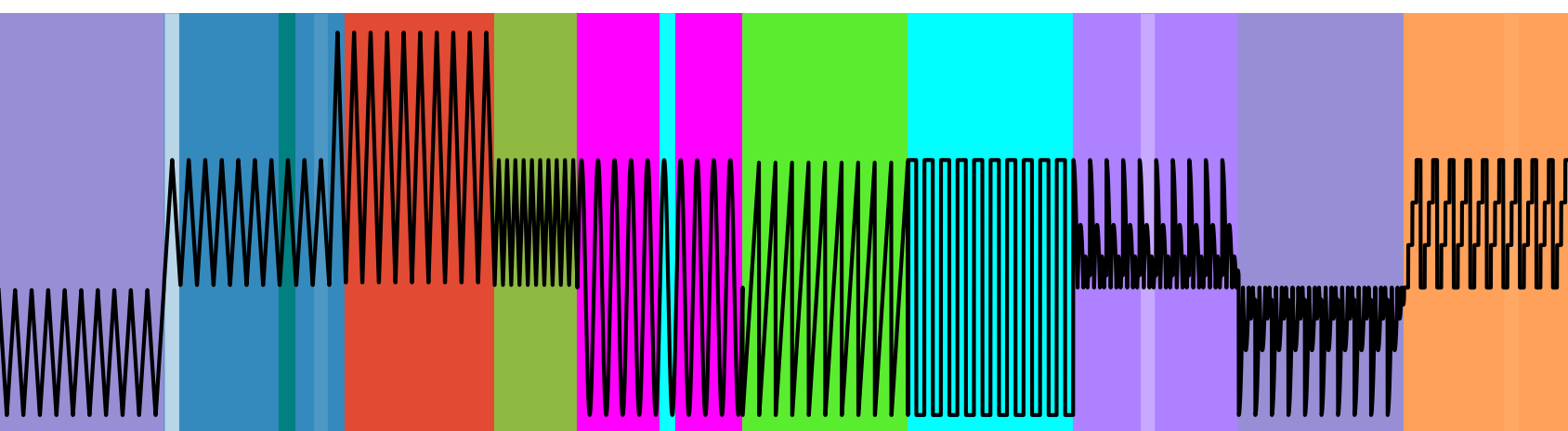}
        \caption{Each subsequence's association with T2P-learned pattern, color mapped.}
    \end{subfigure}
    \hspace{1em} 
    \begin{subfigure}{0.45\textwidth}
        \centering
        \includegraphics[width=\textwidth]{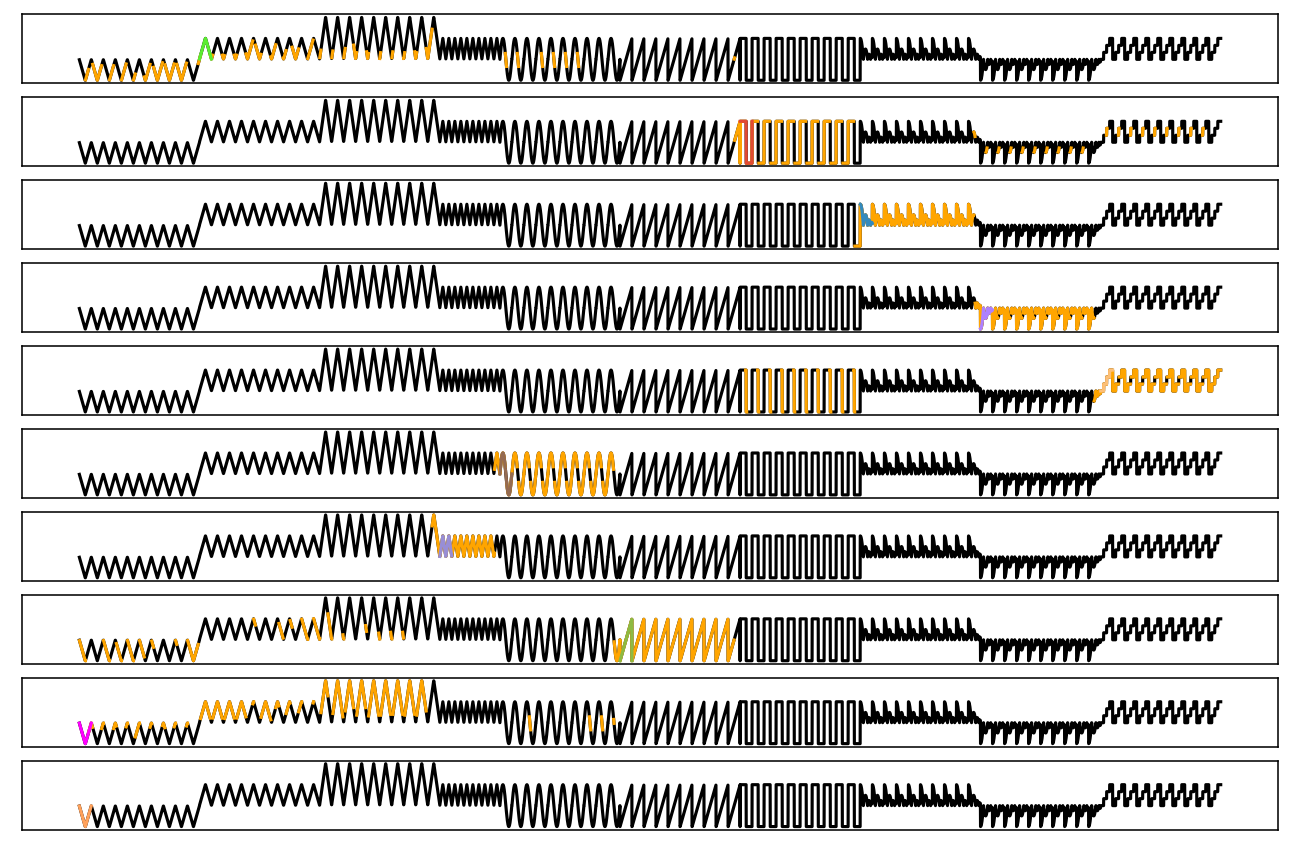}
        \caption{Identified, color-mapped occurrences of MP-Snippet’s discovered patterns depicted in the original time series data. Additional subsequences closely associated with a specific pattern are highlighted in orange.}
        \label{fig:ten_e}
     \end{subfigure}%
    \caption{Visual inspection of T2P and MP-snippets discoveries and identification of pattern occurrences in synthetic data with ten patterns.\vspace{-5mm}}
    \label{fig:ten}
\end{figure}

\begin{table}[h]
 \centering
 \caption{Performance of T2P and MP-Snippets. T2P's performance is averaged over 5 runs (average $\pm$ standard deviation).  MP-Snippets is deterministic, thus the standard deviation is 0. The best-performing algorithm is highlighted in bold for each metric. Dataset SY4-[noise level] consists of synthetic data with four patterns subject to varying noise levels. Dataset SY10 features synthetic data with ten integrated patterns. For the Plane-1 dataset, the F-14's closed and opened positions are treated as separate classes, whereas in the Plane-2 dataset, they are regarded as a single class.}
    \begin{tabular}{cccccc}
    \toprule
    Dataset & Model & Compression & Precision & Recall \\
    \midrule
    SY4-0\% & T2P & \textbf{7.7 $\pm$ 0.6} & \textbf{0.99 $\pm$ 0.01}& \textbf{1.00 $\pm$ 0.00}  \\
        & MP-Snippets  & 1.2 $\pm$ 0.0 & 0.34 $\pm$ 0.00 & 0.75 $\pm$ 0.00  \\
    \midrule
    SY4-30\% & T2P  & \textbf{6.6 $\pm$ 0.8} & \textbf{0.97 $\pm$ 0.01}  & \textbf{1.00 $\pm$ 0.00}  \\
          & MP-Snippets & 1.2 $\pm$ 0.0 & 0.34 $\pm$ 0.00 & 0.50 $\pm$ 0.00  \\
    \midrule
    SY4-70\% & T2P  & \textbf{7.7 $\pm$ 0.6} & \textbf{0.99 $\pm$ 0.01}  & \textbf{1.0 $\pm$ 0.0}  \\
          & MP-Snippets & 1.2 $\pm$ 0.0 & 0.34 $\pm$ 0.00 & 0.50 $\pm$ 0.00  \\
    \midrule
    SY4-100\% & T2P & \textbf{4.8 $\pm$ 0.6} & \textbf{0.90 $\pm$ 0.02}  & \textbf{1.00 $\pm$ 0.00}  \\
           & MP-Snippets & 1.1 $\pm$ 0.0 & 0.28 $\pm$ 0.00 & 0.25 $\pm$ 0.00  \\
    \midrule
    SY10 & T2P & \textbf{8.2 $\pm$ 0.5} & \textbf{0.99 $\pm$ 0.00} & \textbf{1.0 $\pm$ 0.00}  \\
         & MP-Snippets  & 2.3 $\pm$ 0.0 & 0.78 $\pm$ 0.00 & 0.98 $\pm$ 0.00  \\
     \midrule
     Vital Sign & T2P & \textbf{22.8 $\pm$ 0.0} & \textbf{1.00 $\pm$ 0.00} & \textbf{1.00 $\pm$ 0.00}  \\
                & MP-Snippets  & \textbf{22.8 $\pm$ 0.0} & \textbf{1.00 $\pm$ 0.00} & \textbf{1.00 $\pm$ 0.00}  \\
     \midrule
     Audio MNIST & T2P & \textbf{8.3 $\pm$ 2.2} & \textbf{0.96 $\pm$ 0.04} & \textbf{1.00  $\pm$ 0.00}  \\
                 & MP-Snippets  & 0.2 $\pm$ 0.0 & 0.50 $\pm$ 0.00 & 0.50 $\pm$ 0.00 \\
    \midrule
    ECG & T2P & \textbf{3.3 $\pm$ 0.1} & \textbf{0.72 $\pm$ 0.00} & \textbf{1.00  $\pm$ 0.00}  \\
        & MP-Snippets  & 3.0 $\pm$ 0.0 & 0.70 $\pm$ 0.00 & \textbf{1.00 $\pm$ 0.00} \\
    \midrule
     Plane-1 & T2P & \textbf{4.4 $\pm$ 0.2} & \textbf{0.84 $\pm$ 0.00} & {0.98  $\pm$ 0.00}  \\
             & MP-Snippets  & 4.1 $\pm$ 0.0 & 0.82 $\pm$ 0.00 & \textbf{1.00 $\pm$ 0.00} \\
    \midrule
     Plane-2  & T2P & \textbf{10.6 $\pm$ 0.9} & \textbf{0.97 $\pm$ 0.00} & \textbf{1.00  $\pm$ 0.00}  \\
              & MP-Snippets  & 4.2 $\pm$ 0.0 & 0.83 $\pm$ 0.00 & \textbf{1.00 $\pm$ 0.00} \\
    \bottomrule
    \end{tabular}
  \label{tab:comparison}
\end{table}

{\em Multiple pattern discovery.} We also generated synthetic data containing ten distinct patterns, each easily distinguishable by the naked eye. Each pattern is 100 units long and varies in amplitude and frequency. As Figure~\ref{fig:ten} indicates, T2P successfully identified all ten patterns and correctly paired each pattern with the appropriate subsequences in the data. In contrast, MP-Snippets identified only nine out of the ten patterns, with one pattern appearing repetitively. Furthermore, MP-Snippets consistently erred in pairing the first three patterns with the correct subsequences, as demonstrated in Table~\ref{tab:comparison}. In Figure~\ref{fig:ten}d, the colors indicate the pattern to which the subsequence belongs and the color intensity reflects the maximum corresponding $z$ value. In the MP-Snippets visualization shown in Figure~\ref{fig:ten}e, one occurrence of the pattern is highlighted in the pattern color, while the remaining close matches to that pattern are shown in orange.


\subsection{Vital Sign Data}

In this experiment, we evaluate summarization of ICU vital sign data. In this setting, a clinician visits the patient's bedside hourly to collect vitals ~\cite{drews2008patient, forde2014intentional}. Events such as sneezing or staff presence can alter intrathoracic pressure and impact vital sign readings. This dataset contains known patterns that occur while physically changing patient orientation on a sensor-equipped bed. 
We include this dataset to illustrate that for simple time series, T2P performs similarly to earlier methods like MP-Snippets (see Table~\ref{tab:comparison}). For this time series, T2P and Snippets both find the embedded patterns and achieve a Compression of 22.8, precision of 1.0, and recall of 1.0. T2P did incur a smaller computation cost, requiring 19.75 seconds in comparison with the MP-Snippets run time of 265.04 seconds.

\subsection{Audio Data}

In the next experiment, we analyze the ability of T2P and MP-Snippets to discover effective summarizing patterns in more complex audio waveform data. For this experiment, we utilize data from the AudioMNIST dataset~\cite{becker2018interpreting}. 
This data contains eleven spoken occurrences of the number ``one'' followed by eleven occurrences of the number ``two'', collected from a 30-year-old male subject with a German accent. The recordings were downsampled to 8kHz and zero-padded to generate an 8000-dimensional vector per recording.

We are interested in determining whether the algorithms can discover summarizing patterns, defined by their ability to compress the original time series. Given the two spoken numbers, we also calculate precision and recall based on the ability to identify patterns for these two distinct sounds.
As indicated in Table~\ref{tab:comparison}, the T2P patterns provided much greater data compression than those discovered by MP-Snippets. MP-Snippets identified two patterns, though both originated from sequences where the subject articulated the number {\it one}. However, it failed to discover a pattern corresponding to the number {\it two}. In contrast, T2P learned distinct patterns for each spoken numeral and associated each pattern with its corresponding subsequences in the data. The precision and recall scores listed in Table \ref{table:comparison} provide additional evidence that T2P was able to more effectively find correct summarizing patterns in the complex timeseries.
Another point of interest is the substantial difference in runtime between the two algorithms. T2P required 607.73 seconds, while MP-Snippets required 15,237.27 seconds to complete.

\subsection{ECG Data}

Next, we analyze the ability of T2P and MP-Snippets to discover summarizing patterns in ECG waveform data~\cite{olszewski2001generalized}. Each waveform traces the electrical activity recorded during one cardiac cycle (i.e., heartbeat) by a single electrode. This represents data recorded during a normal heartbeat and during a heartbeat exhibiting behavior indicative of a cardiac condition called myocardial infarction. As indicated in Table~\ref{tab:comparison}, the T2P patterns provide much greater data compression than those discovered by MP-Snippets. Although recall is optimal for both algorithms, T2P also results in higher precision than T2P. The results indicate again that T2P is more effective at summarizing complex patterns than the predecessor method. T2P required 74.48 seconds for this case, while MP-Snippets required 15.16 seconds.
\begin{wrapfigure}{r}{0.45\textwidth}
    \begin{center}
    \includegraphics[width=0.5\textwidth]{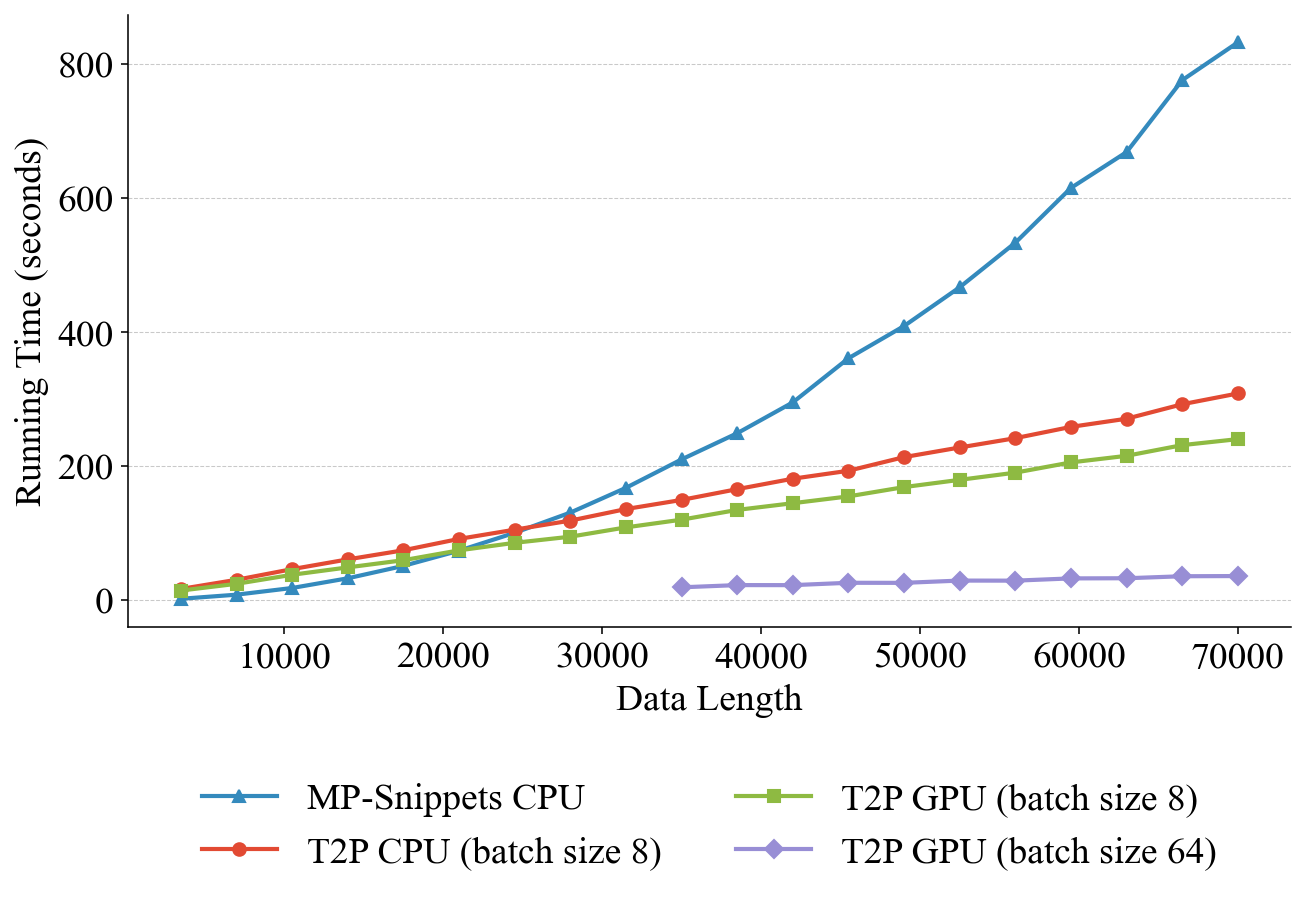}
    \end{center}
    \caption{Run times for MP-Snippets (CPU) and T2P (CPU, GPU with batch size 8, GPU with batch size 64) as a function of data size, which is capped at 70,000 due to the inability of MP-Snippets to finish on larger datasets.}
    \vspace{0pt}
    \label{fig:scalability}
\end{wrapfigure}

\subsection{Plane Data}

The plane dataset~\cite{thakoor2005hidden} was created by capturing digital images of airplane die-cast replica models.
The pictures were then converted to outlines and transformed into pseudo time series. One of the airplanes, the F-14, is represented pictorially by two distinct shapes: one with its wings closed and another with its wings open. However, 
the difference between these positions is subtle. T2P discerns a single generalized pattern for the F-14. This outcome is a manifestation of T2P's objective to maximally compress the data, sometimes incurring information loss.
While MP-Snippets managed to identify this slight difference, it incorrectly conflated the distinctly-shaped F-15 with the F-14. While T2P did not distinguish the two positions of the F14, it yielded a higher compression rate, as shown in Table \ref{tab:comparison}.
\begin{wrapfigure}{r}{0.45\textwidth}
    \begin{center}
        \includegraphics[width=.5\textwidth]{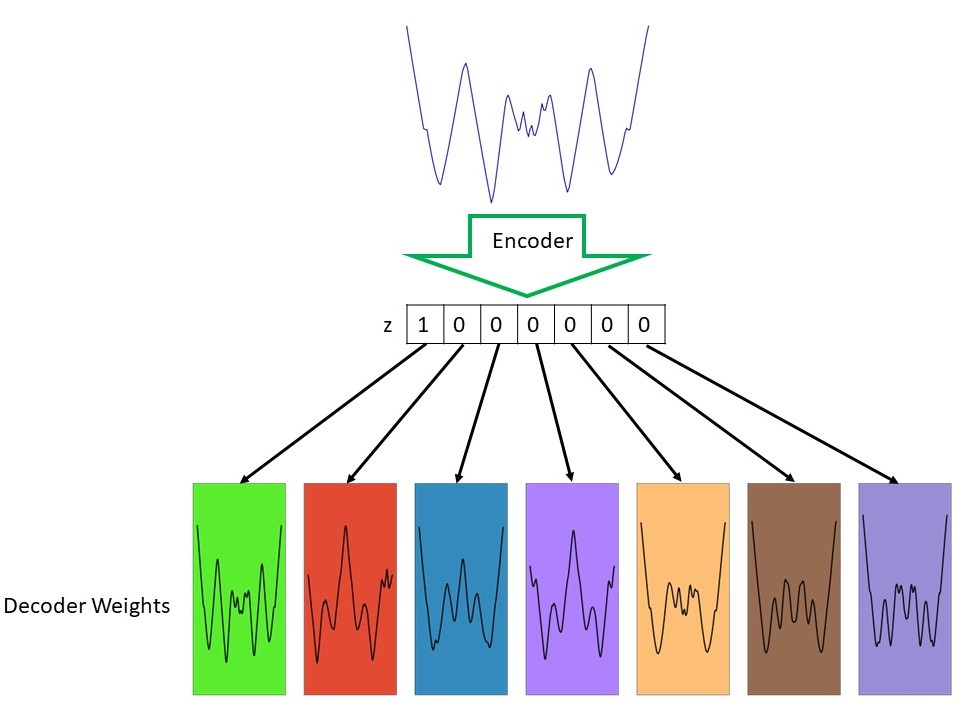}
    \end{center}
    \caption{When the vector generated by the T2P encoder is multiplied by the weights of the learned decoder kernels, the vector $z =[1,0,0,0,0,0,0]$ indicates that only the first kernel (the first learned pattern) is activated.}
    \vspace{-40pt}
    \label{fig:t2platent}
\end{wrapfigure}
\subsection{Scalability}

Because prior time series summarization algorithms rely on exhaustive search, they face difficulties in scaling to large datasets. In this experiment, we examine the performance of MP-Snippets and T2P on the synthetic dataset shown in Figure~\ref{fig:mechanism}, repeated to yield time series of varying sizes.

As shown in Figure~\ref{fig:scalability}, MP-Snippets exhibits a superlinear run time, while T2P demonstrates near-linear scalability. Furthermore, the T2P algorithm can leverage GPU resources and adjust its batch size based on the data size to improve performance. When harnessing this capability, T2P maintains an almost constant run time as the time series grows.


We also observe that even when the patterns are repeated more often to increase dataset size, MP-Snippets still does not discover the last pattern, as described in Section~\ref{section:synthetic}.
This omission is attributed to the inherent bias of the algorithm, due to its reliance on a Euclidean distance similarity function, which might not be the optimal choice for datasets such as these.

\subsection{Interpretable Latent Space}

VAEs use a latent space to learn compressed, abstract data representations. Within this context, the concept of a sparse latent space becomes particularly intriguing. When we reference a ``sparse latent space'' in the VAE paradigm, this implies that for a given input, only a limited number of dimensions (or latent variables) deviate considerably from zero. In contrast, a dense representation would entail that a majority, if not all, of the dimensions hold significant values.

Focusing on the T2P model, the sparse nature of its latent space lends itself to enhanced interpretability. The sparse latent space of T2P provides clear indicators about which specific kernels in the decoder (patterns) are active and vital for reconstructing the input data. This activation can be perceived as a spotlight, highlighting those kernels that play pivotal roles in the data reconstruction process, as illustrated in Figure \ref{fig:t2platent}.

\section{Discussion and Conclusions}
In this study, we introduced an innovative method for time series summarization using an information-theoretic VAE model. The foundation of our approach lies in compression and the identification of dominant patterns that facilitate data compression. A clear advantage of our method, as evidenced in the Plane dataset, is its proficiency in data compression. However, a limitation arises when two patterns are highly similar; our current model might not effectively differentiate between them. Consequently, T2P might not be suitable for such instances. At present, T2P accommodates univariate time series data. While it is feasible to feed each axis of a multivariate signal into the model separately, T2P does not directly accept multivariate data. Addressing this limitation is a future objective. Moreover, alternative architectural designs merit exploration to assess their impacts and potential enhancements.

Our results indicate that T2P surpasses existing techniques regarding compression, precision, recall, and scalability. Yet, a current constraint is that the algorithm mandates user-specified pattern lengths and pattern counts. Future iterations aim to mitigate this requirement. By identifying fewer patterns, we reduce pattern overlap, albeit at the potential expense of overlooking certain compressive patterns. Quantifying these factors will inform an unsupervised strategy to fine-tune such determinations.

Our research emphasizes time series summarization via MDL-driven pattern detection. This approach holds substantial implications by enabling users to decipher intricate time series data and recognize foundational patterns with increased efficiency. Such capabilities are especially beneficial in areas like healthcare, speech analysis, and signal processing.


\section*{Acknowledgments}
We would like to acknowledge support for this project
from the National Science Foundation (NSF grant IIS-9988642)
and the Multidisciplinary Research Program of the Department
of Defense (MURI N00014-00-1-0637). 


\newpage

\appendix

\section{Experimental Conditions}

\subsection{Implementation}

We implemented the T2P model using the PyTorch framework. All code is available online at \url{https://github.com/alirezaghods/T2P-Time-to-Pattern/blob/main/Appendix.pdf}. For MP-Snippets, we utilized the library furnished by the Matrix Profile Foundation \cite{MatrixProfile2023}.

\subsection{T2P Architecture} \label{appendix:t2p}

The T2P architecture remained fixed for all reported experiments. We utilize the Adam optimizer \cite{kingma2014adam}. Detailed information about the networks used and the dimensions of inputs and outputs at different stages is provided in Table \ref{tab:architecture}.

\begin{table}[ht]
\caption{T2P network architecture details.}
\label{sample-table}
\center
\begin{tabular}{l c c c c c } \hline
Operation & Kernel & Feature map & Padding & Stride & Nonlinearity \\ \hline
\multicolumn{6}{c}{Input: Batch of time series subsequence of length $m$} \\ \hline
1D Conv & 3 & 12 & 0 & 1 & ReLU \\
1D Conv & 3 & 24 & 0 & 1 & ReLU \\
1D Max-Pool & 2 & - & 0 & 2 & - \\
1D Conv & 3 & 32 & 0 & 1 & ReLU \\ \hline
\multicolumn{6}{c}{Output: Batch of time series subsequence of length $\tilde{m} = (((m - 3 + 1 - 3 + 1)/2 - 3 + 1))$}  \\
\multicolumn{6}{c}{Input: Batch of time series subsequence of length $\tilde{m}$} \\ \hline
3D Conv & $n_{patterns} \times \tilde{m}$ & $2$ & $(n_{patterns} - 1) \times 0 \times 0$ & 1 & SoftPlus \\ \hline
\multicolumn{6}{c}{Output: $\text{batch size} \times 1 \times n_{patterns} \times 1 \times 1$} \\
\multicolumn{6}{c}{Input: $\text{batch size} \times 1 \times n_{patterns} \times 1 \times 1$} \\ \hline
3D T. Conv & $n_{patterns} \times m \time 1$ & 1 & $(n_{patterns} - 1) \times 0 \times 0$ & 1 & - \\ \hline
\multicolumn{6}{c}{Output: Batch of time series subsequence of length $m$} \\ \hline
\end{tabular}
\label{tab:architecture}
\end{table}

\paragraph{Encoder}
The encoder network commences with a series of convolutional layers. Each layer uses 1D filters to analyze subsequences and extract key features. The resulting feature maps are passed through a final convolutional layer equipped with 3D filters, designed to match the number of patterns. We selected a CNN based on previous research~\cite{ismail2019deep} demonstrating its superior performance on time series data compared to other architectures. Future work can investigate the impact of other model structures.

\paragraph{Decoder} The decoder's goal is to learn kernels for the purpose of reconstructing the data. Consequently, we chose a deconvolution layer.  The decoder is fed by the encoder output, {\bf z}, and is composed of a single deconvolution layer equipped with $n_{patterns}$ filters. These deconvolution filters contain the patterns of interest.

\subsection{Model hyperparameters} \label{model_hyperparameters} In the T2P model, we tune three hyperparameters: $a$, $\lambda_1$, and $\lambda_2$. As detailed in Section 4, $a$ designates the location of the prior distribution. We suggest setting $a$ within the range of $0.8 \leq a \leq 0.9$, as this range promotes sparsity in the latent space and encourages pattern diversity.

The subsequent step involves the fine-tuning of parameters $\lambda_1$ and $\lambda_2$, aiming to enhance pattern fidelity. Our experimental investigations have shown that the best outcomes are achieved when $\lambda_1$ and $\lambda_2$ are selected within the ranges of $0.7 \leq \lambda_1 \leq 0.9$ and $0.2 \leq \lambda_2 \leq 0.3$, respectively.
Table \ref{tab:hyperparam} provides a detailed summary of the hyperparameters set for each conducted experiment. 

\begin{table}[h]
    \caption{Parameters utilized for the experiments detailed in this paper. Here, $a$ represents the location of the BinConcrete prior, while $\lambda_1$ and $\lambda_2$ denote the temperatures for the relaxed approximate posterior and prior distributions, respectively. $n_{patterns}$ indicates the number of patterns. ``Pattern length'' refers to the length of the patterns the model is trained on and capable of processing. Please note: The same hyperparameters were employed across all synthetic data sets, irrespective of noise levels, hence only one set is reported.}
    \label{tab:hyperparam}
    \center
    \begin{tabular}{l c c c c c c c} \hline
        Datasets & a & $\lambda_1$ & $\lambda_2$ & $n_{patterns}$ & pattern length & epochs & learning rate\\ \hline
        SY4 & 0.8 & 0.83 & 0.21 & 4 & 100 & 1000 & 1e-3 \\
        SY10 & 0.8 & 0.83 & 0.21 & 4 & 100 & 1000 & 1e-3 \\
        Vital Sign & 0.8 & 0.83 & 0.23 & 2 & 850 & 1000 & 1e-3 \\
        Audio MNIST & 0.8 & 0.83 & 0.23 & 2 & 8000 & 2000 & 1e-4 \\
        ECG & 0.8 & 0.91 & 0.25 & 2 & 96 & 2000 & 1e-3 \\
        Plane & 0.92 & 0.90 & 0.10 & 7 & 144 & 4000 & 1e-3 \\ \hline
    \end{tabular}
\end{table}

\section{Dealing with Noisy Conditions in Time Series Pattern Analysis} \label{app:dealingNoise}

In the real world, time series data often comes with a certain level of noise due to various sources such as measurement error. A good time series summarization method should be robust against this noise, being able to ignore irrelevant fluctuations and highlight true underlying patterns. A model that can handle noise effectively can better generalize to unseen data. A well-known stochastic process in time series analysis - the Autoregressive (AR) process of order 1, AR(1) - offers a useful framework for understanding this. This process is described by the equation:

\begin{equation}
X_t = \alpha X_{t-1} + \epsilon_t
\end{equation}

where $X_t$ denotes our time series data at time $t$, $\alpha$ is a coefficient that captures the influence of the preceding time point ($X_{t-1}$) on the current one ($X_t$), and $\epsilon_t$ signifies the error term or noise at time $t$.

Under standard circumstances, $\epsilon_t$ is often treated as Gaussian white noise, i.e., a sequence of random variables that are identically distributed (with each drawn from a normal distribution), independent, and with a mean of zero. Such noise is inherently unpredictable.

Nevertheless, in various real-world scenarios, the noise term $\epsilon_t$ might not be purely random but could be shaped by other latent variables or processes. For instance, $\epsilon_t$ could follow a distribution affected by unobserved factors, which can be represented as $\epsilon_t = f(p, q)$. Here, $f$ denotes a function encapsulating the influence of latent variables $p$ and $q$ on the noise $\epsilon_t$.

When the entire equation $X_t = \alpha X_{t-1} + f(p, q)$ is considered, it is evident that the noise is not purely random, but rather, is influenced by underlying latent variables. Thus, an efficacious time series pattern recognition method should not solely focus on the principal process ($X_t = \alpha X_{t-1}$) but also comprehend the inherent noise structure ($\epsilon_t = f(p, q)$) within the data. This is precisely the capability that our T2P model seeks to deliver.

Traditional search-based methods, which largely depend on similarity functions like the Euclidean distance, often struggle to capture this nuanced understanding of noise. Conversely, our T2P model utilizes a learning-based strategy aimed at discerning patterns that account for a significant portion of the time series data. As a result, it works towards minimizing the mean square error between the input subsequence and the sequence recreated using the model's kernels, thereby dealing more effectively with the 'noise' within the data. To demonstrate this difference in approach and effectiveness, we have executed both T2P and MP-Snippets on data integrated with Gaussian white noise.

The capability of T2P to discern patterns amidst the noise is a direct consequence of its learning-based strategy. By actively learning the primary components of the time series data and the inherent noise structure, T2P is able to minimize the mean square error between the input subsequence and the sequence reconstructed using the model's kernels. This is a critical aspect of time series pattern analysis, as the intricacies of the data are better understood, and more accurate patterns are discerned even in noisy environments.

In contrast, MP-Snippets, which leans heavily on search-based methods, reveals limitations under noisy conditions. Such methods, which largely rely on similarity functions like the Euclidean distance, are inherently limited in their ability to comprehend the underlying noise structure. As a result, their performance tends to deteriorate in the presence of significant noise. This suggests that these techniques may not be sufficiently robust for real-world applications where noise is a prevalent factor.

This analysis underscores the importance of using learning-based strategies, like the one employed by the T2P model, in time series pattern analysis. In real-world scenarios, where the noise may be influenced by various latent factors, understanding and accounting for the noise structure becomes an essential part of the analysis. This evaluation demonstrates the superiority of our T2P model in this regard, setting a compelling precedent for future advancements in time series pattern analysis.

\section{Hyperparameter Settings} \label{app:parameter_setting}

Our empirical analysis indicates that most parameters perform well using the default settings across all tested datasets. The primary parameters that need adjustment are the number of patterns and pattern length. As demonstrated in our experiments, T2P is fairly robust when varying hyperparameters. However, these choices can impact results and we discuss the most significant choices here.

\paragraph{Guiding Hyperparameter Selection}

The T2P model is fundamentally designed to construct a sparse latent space. Sparsity is necessary because it enhances model interpretability and the ability to capture abstract features by learning a disentangled representation of data.This underscores the importance of sparsity as a critical performance indicator. Increasing sparsity not only shows that the model is learning a sparse space but also suggests that the model is succeeding in capturing a set of distinct patterns. In an ideal scenario, for every input subsequence, a single kernel should suffice to reproduce the input, negating the need for a group of kernels.

Sparsity is quantified with reference to Hoyer's measure~\cite{hurley2009comparing}. Given a vector \(y \in \mathbb{R}^d\), the measure is defined as:

\begin{equation}
S(y) = \frac{\sqrt{d} - \frac{\|y\|_1}{\|y\|_2}}{\sqrt{d} - 1} \in [0, 1]
\end{equation}

The measure ranges from 0 for a dense vector to 1 for a sparse vector. In a similar vein to previous work~\cite{higgins2017beta}, we do not directly apply this measure to the mean encoding of each datapoint. Instead, we first normalize each dimension to have a standard deviation of 1 under its aggregate distribution, represented as \(\bar{z}_d = \frac{z_d}{\sigma(z_d)}\), where \(\sigma(z_d)\) denotes the standard deviation of dimension \(d\) of the latent encoding over the dataset. This normalization is critical to prevent different dimensions from being sparse simply because they vary along different length scales. The overall sparsity is then calculated by averaging at each epoch:

\begin{equation}
\text{Sparsity} = \frac{1}{n} \sum_{i=1}^{n} S(\bar{z}_i)
\end{equation}

\begin{figure}
    \centering
    \centering
    \begin{subfigure}{0.45\textwidth}
        \centering
        \includegraphics[width=6cm]{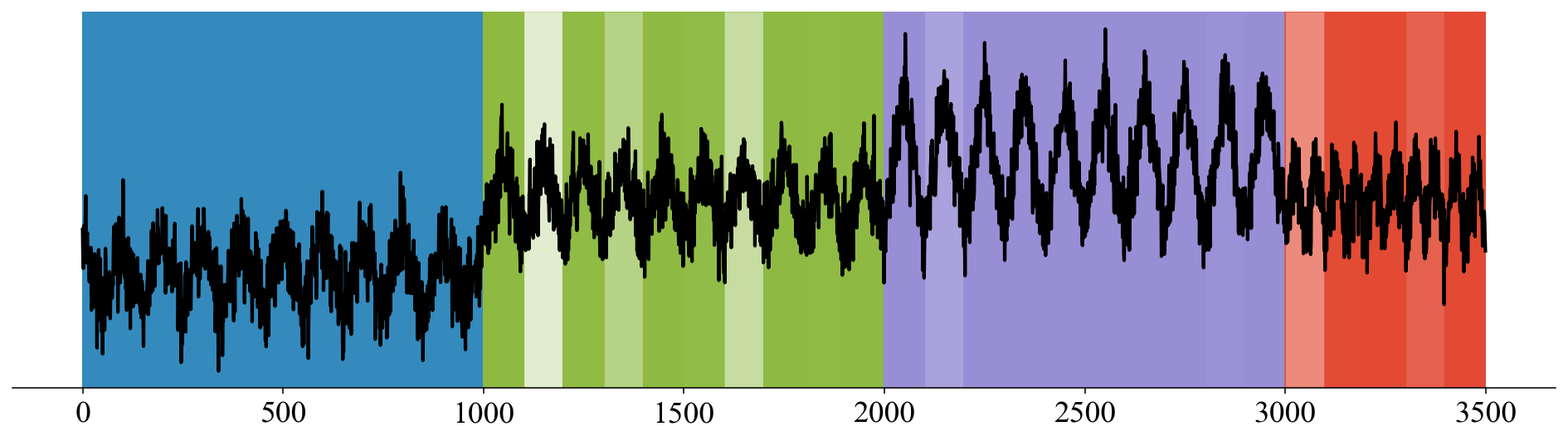}
        \caption*{Subsequence associations with patterns.}
        \includegraphics[height=2.2cm]{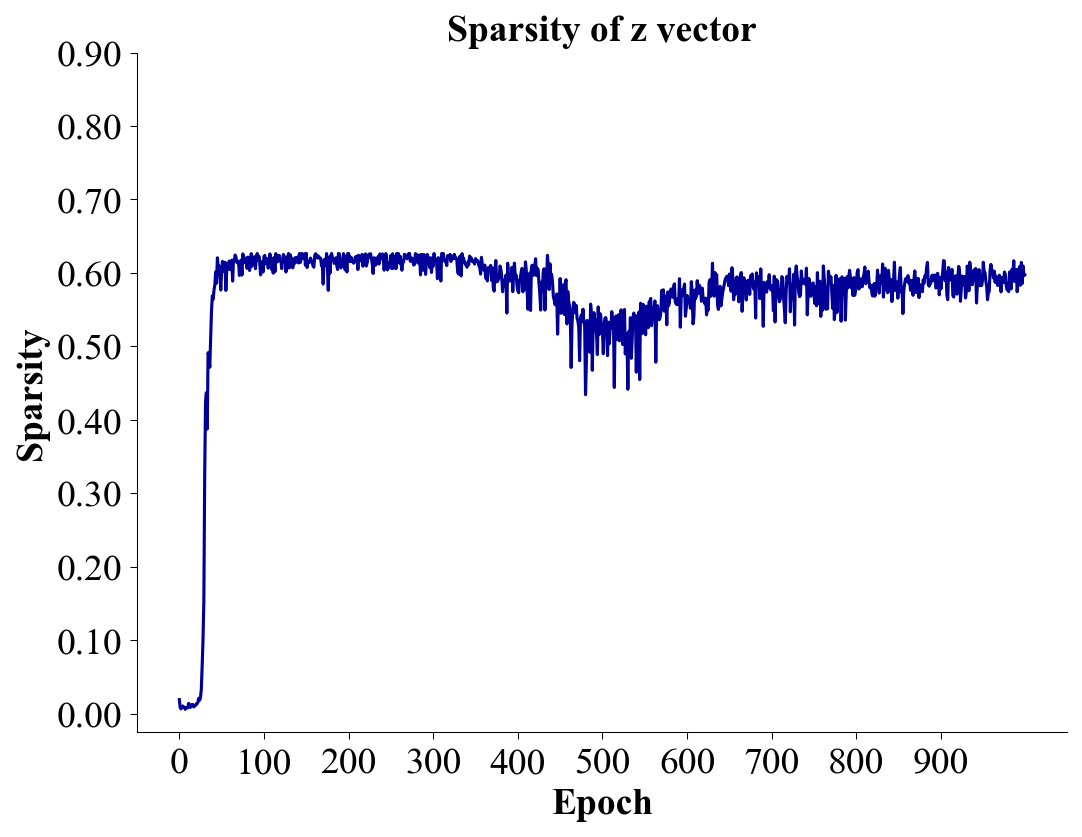}
        \caption*{sparsity}
        \caption{Training parameters: $\lambda_1=0.83$,$\lambda_2=0.21$, \\ $a=0.8$,$n_{patterns}=4$,$pattern length = 100$}
        \label{fig:sparcity_best}
    \end{subfigure}
    \hspace{.1em} 
    \begin{subfigure}{0.45\textwidth}
        \centering
        \includegraphics[width=6cm]{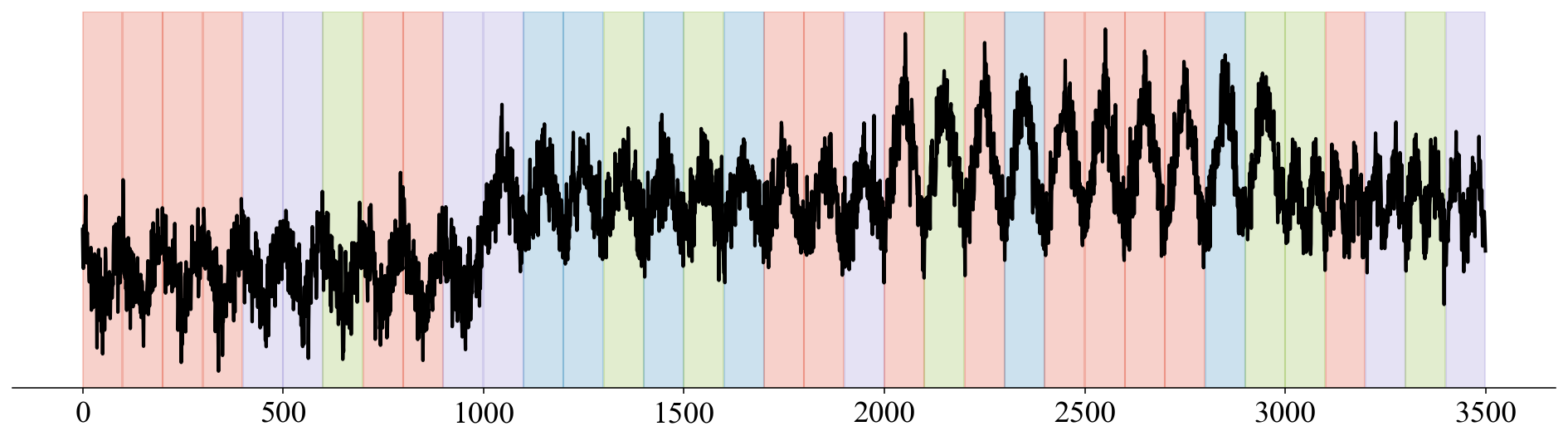}
        \caption*{Subsequence associations with patterns.}
        \includegraphics[height=2.2cm]{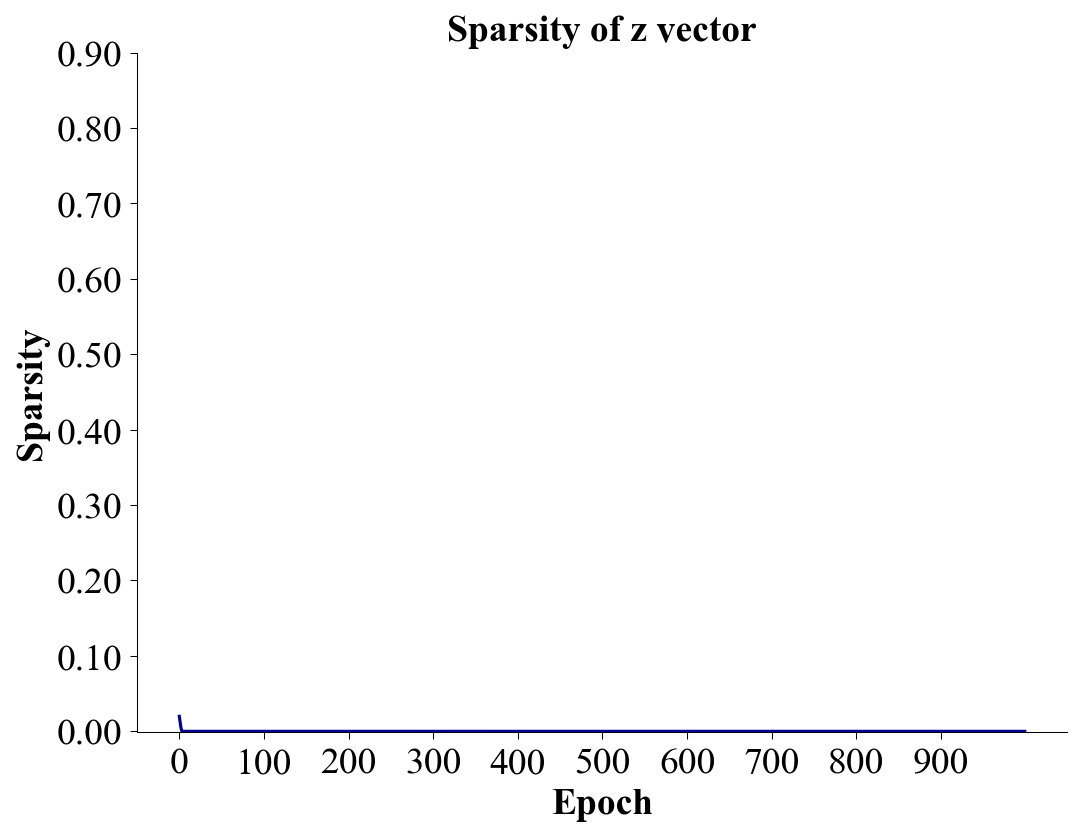}
        \caption*{sparsity}
        \caption{Training parameters: $\lambda_1=0.83$,$\lambda_2=0.21$, \\ $a=0.3$,$n_{patterns}=4$,$pattern length = 100$}
        \label{fig:sparcity_a}
    \end{subfigure}
    \vspace{1em}
    \begin{subfigure}{0.45\textwidth}
        \centering
        \includegraphics[width=6cm]{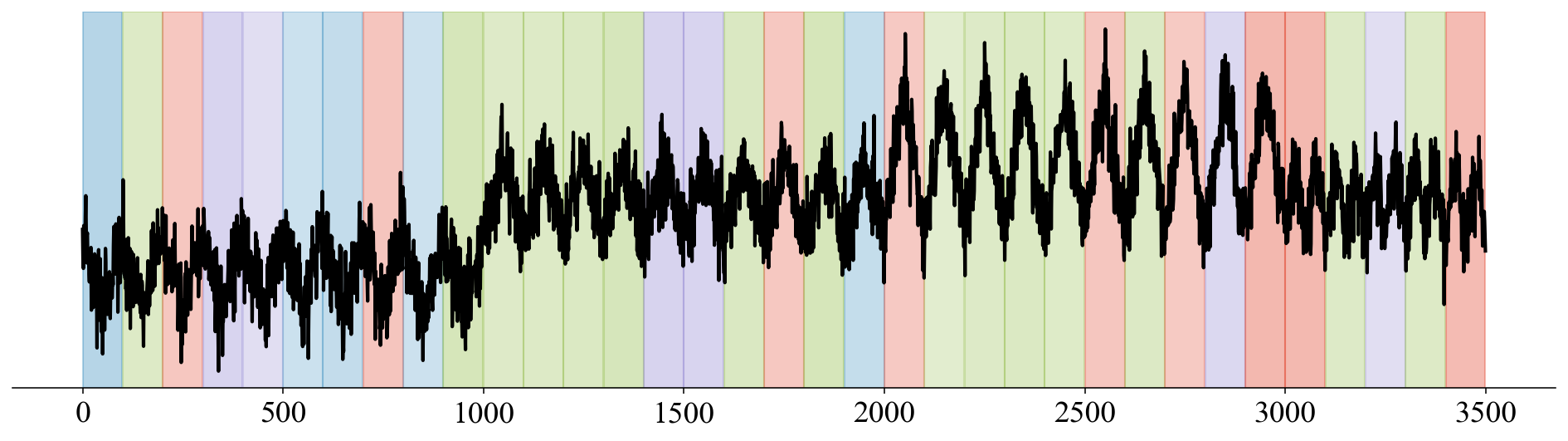}
        \caption*{Subsequence associations with patterns.}
        \includegraphics[height=2.2cm]{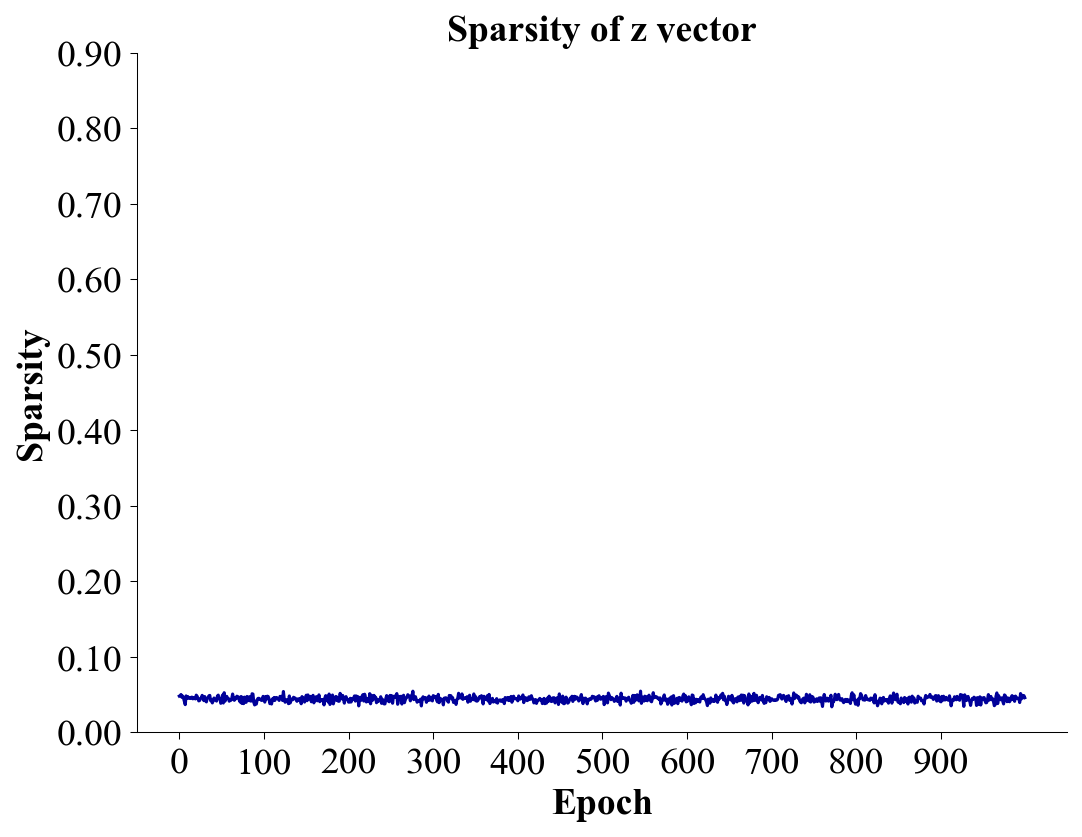}
        \caption*{sparsity}
        \caption{Training parameters: $\lambda_1=0.23$,$\lambda_2=0.81$, \\ $a=0.8$,$n_{patterns}=4$,$pattern length = 100$}
        \label{fig:sparcity_lambda}
    \end{subfigure}
    \hspace{0.1em}
    \begin{subfigure}{0.45\textwidth}
        \centering
        \includegraphics[width=6cm]{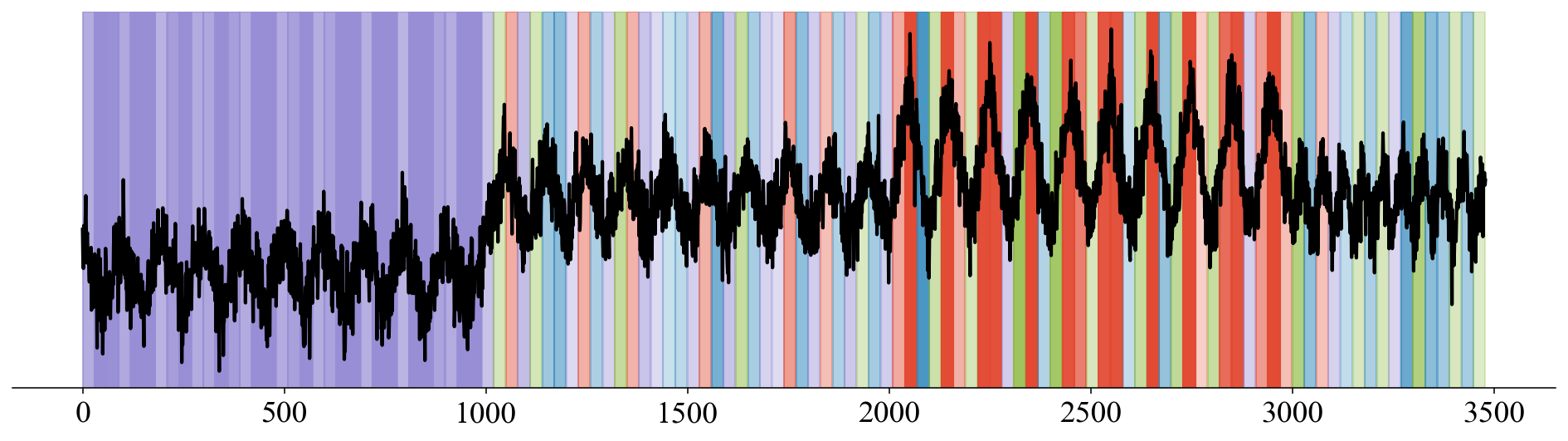}
        \caption*{Subsequence associations with patterns.}
        \includegraphics[height=2.2cm]{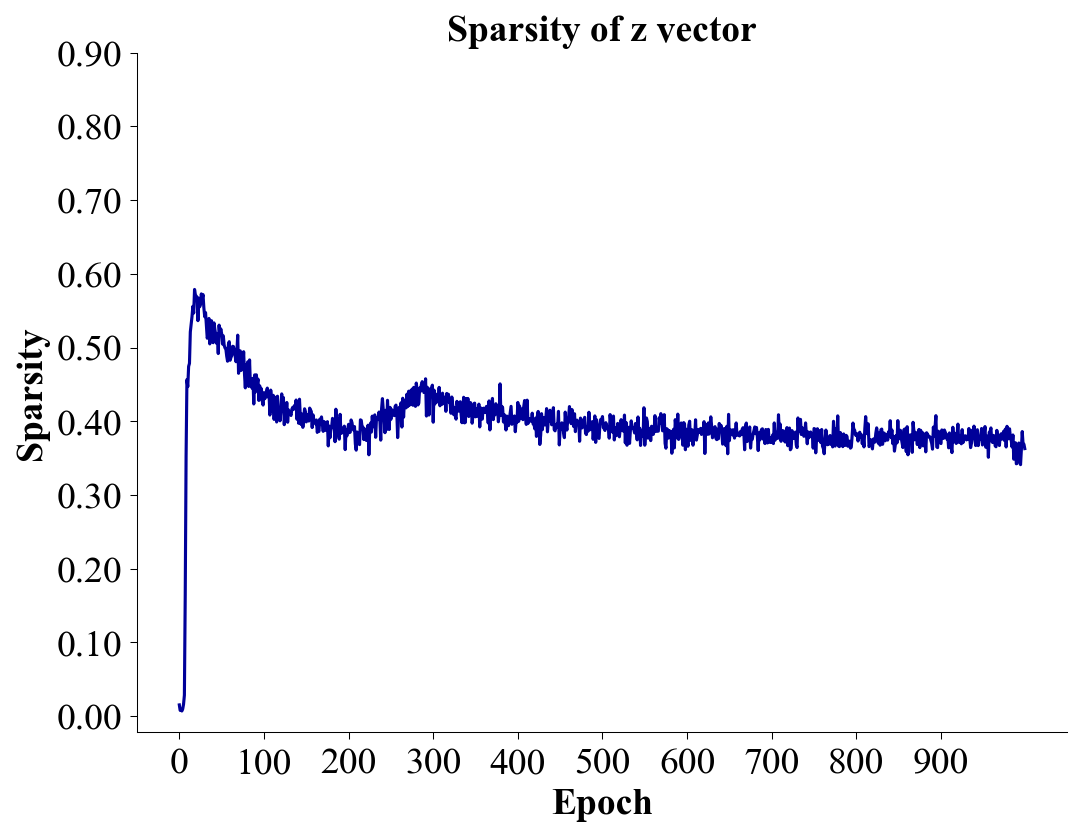}
        \caption*{sparsity}
        \caption{Training parameters: $\lambda_1=0.83$,$\lambda_2=0.21$, \\ $a=0.8$,$n_{patterns}=4$,$pattern length = 30$}
        \label{fig:sparcity_length}
    \end{subfigure}
    \vspace{0.1em}
    \begin{subfigure}{0.45\textwidth}
        \centering
        \includegraphics[width=6cm]{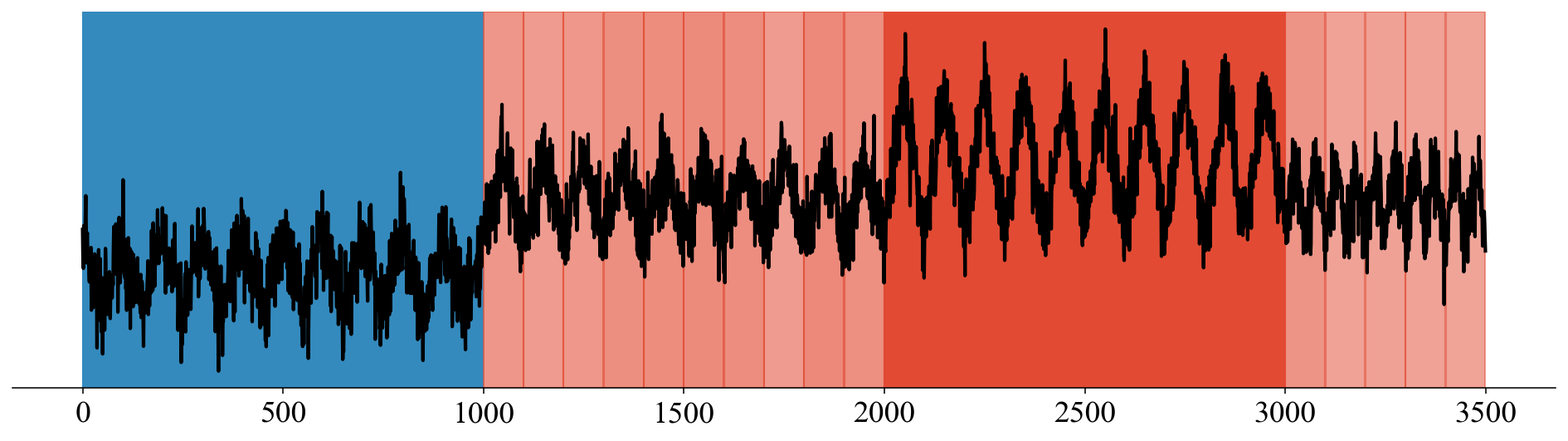}
        \caption*{Subsequence associations with patterns}
        \includegraphics[height=2.2cm]{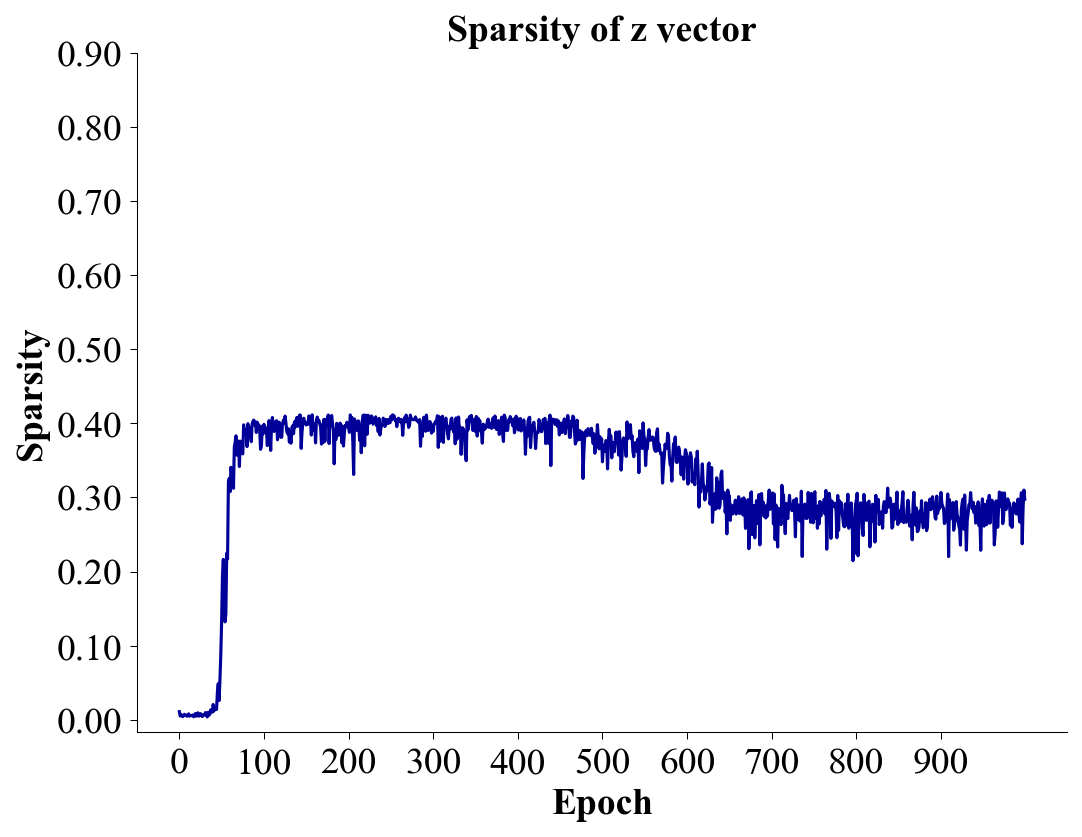}
        \caption*{sparsity}
        \caption{Training parameters: $\lambda_1=0.83$,$\lambda_2=0.21$, \\ $a=0.8$,$n_{patterns}=2$,$pattern length = 100$}
        \label{fig:sparcity_num}
    \end{subfigure}
    \caption{Effect of hyperparameters on latent space sparsity.}
    \label{fig:sparsity}
\end{figure}

At the onset, sparsity is low, then rises as the model trains. Figure \ref{fig:sparcity_best} displays the model trained with optimally tuned hyperparameters. As depicted, sparsity increases to its highest value, $\approx 0.6$. The location of the BinConcrete distribution impacts the ability to achieve a sparse latent space. If this is set too low, as illustrated in Figure \ref{fig:sparcity_a}, T2P fails to learn a sparse latent space and thus, the sparsity remains near zero.

The sparsity of the model is also influenced by parameters $\lambda_1$ and $\lambda_2$. Figure \ref{fig:sparcity_lambda} demonstrates that selecting these values impacts the ability of the model to learn a diverse set of patterns. Two additional crucial parameters to consider are pattern length and the number of patterns. When we underestimated the pattern length, as shown in Figure \ref{fig:sparcity_length}, T2P was still able to learn the embedded patterns, but it did not achieve the highest sparsity ($\approx 0.4$). Similarly, when we underestimated the number of patterns, T2P again did not achieve the highest sparsity ($\approx 0.3$), as shown in Figure \ref{fig:sparcity_num}.

If the sparsity does not display an increasing trend, it may suggest that the model is not effectively learning a sparse space, as illustrated in Figure \ref{fig:sparsity}. Under these circumstances, it might be necessary to halt the training process and adjust the hyperparameters. Because \(\lambda_1\) and \(\lambda_2\) have a significant impact on pattern fidelity, these parameters should be selected first, after which pattern length can be adjusted if the patterns remain unclear.

\begin{figure}
    \centering
    \includegraphics[width=0.4\textwidth]{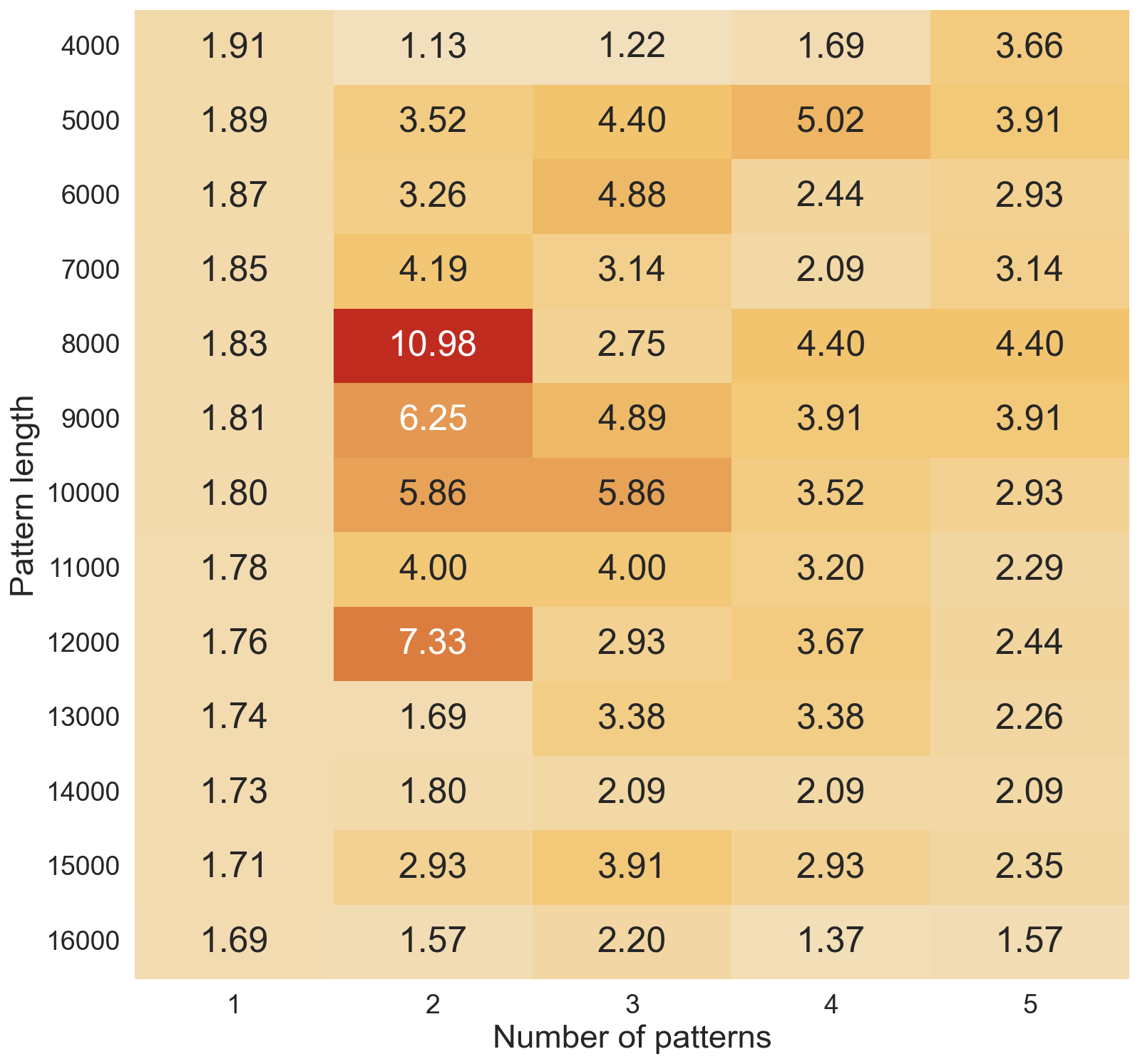}
    \caption{The heat map depicts the compression rates (with higher rates being more desirable) for a range of pattern lengths and number of patterns applied to audio data.}
    \label{fig:ch8:heatmap:coice}
\end{figure}

As previously discussed, T2P's training hinges on two crucial parameters: the expected quantity of patterns and the pattern length. Because T2P is designed to optimize summarizability, or compression, of the raw data, we can use this criteria to automate hyperparameter selection. To exemplify this idea, we trained T2P using audio data with varying pattern cardinality and length. In particular, we considered the number of patterns in the range 1--5. We also varied pattern length in the range 4000--1600. Here, we considered increments of 1000. Bayesian optimization could be used to explore a larger space of choices more efficiently. In this instance, the optimal combination was two patterns with a length of 8000, and the second best combination was also two patterns, but with a pattern length of 12000, as illustrated in Figure \ref{fig:ch8:heatmap:coice}.

\bibliographystyle{unsrt}  
\bibliography{sample}  

\begin{thebibliography}{10}

\bibitem{byabazaire2020data}
John Byabazaire, Gregory O'Hare, and Declan Delaney.
\newblock Data quality and trust: A perception from shared data in {IoT}.
\newblock In {\em Proceedings of the IEEE International Conference on
  Communications Workshops}, pages 1--6. IEEE, 2020.

\bibitem{ahmed2019data}
Mohiuddin Ahmed.
\newblock Data summarization: a survey.
\newblock {\em Knowledge and Information Systems}, 58(2):249--273, 2019.

\bibitem{el2014interpretable}
Kareem El~Gebaly, Parag Agrawal, Lukasz Golab, Flip Korn, and Divesh
  Srivastava.
\newblock Interpretable and informative explanations of outcomes.
\newblock {\em Proceedings of the VLDB Endowment}, 8(1):61--72, 2014.

\bibitem{wen2018interactive}
Yuhao Wen, Xiaodan Zhu, Sudeepa Roy, and Jun Yang.
\newblock Interactive summarization and exploration of top aggregate query
  answers.
\newblock In {\em Proceedings of the VLDB Endowment}, volume~11, pages
  2196--2208, 2018.

\bibitem{youngmann2022guided}
Brit Youngmann, Sihem Amer-Yahia, and Aurelien Personnaz.
\newblock Guided exploration of data summaries.
\newblock {\em arXiv preprint arXiv:2205.13956}, 2022.

\bibitem{yu2009takes}
Cong Yu, Laks Lakshmanan, and Sihem Amer-Yahia.
\newblock It takes variety to make a world: diversification in recommender
  systems.
\newblock In {\em Proceedings of the International Conference on Extending
  Database Technology: Advances in Database Technology}, pages 368--378, 2009.

\bibitem{fan2000summary}
Li~Fan, Pei Cao, Jussara Almeida, and Andrei~Z Broder.
\newblock Summary cache: a scalable wide-area web cache sharing protocol.
\newblock {\em IEEE/ACM Transactions on Networking}, 8(3):281--293, 2000.

\bibitem{hooi2017b}
Bryan Hooi, Shenghua Liu, Asim Smailagic, and Christos Faloutsos.
\newblock {BeatLex:} summarizing and forecasting time series with patterns.
\newblock In {\em Proceedings of the Machine Learning and Knowledge Discovery
  in Databases: European Conference}, pages 3--19. Springer, 2017.

\bibitem{Imani2020IntroducingTS}
Shima Imani, Frank Madrid, Wei Ding, Scott~E. Crouter, and Eamonn~J. Keogh.
\newblock Introducing time series snippets: a new primitive for summarizing
  long time series.
\newblock {\em Data Mining and Knowledge Discovery}, 34:1713 -- 1743, 2020.

\bibitem{plessen2020integrated}
Mogens~Graf Plessen.
\newblock Integrated time series summarization and prediction algorithm and its
  application to {COVID-19} data mining.
\newblock In {\em Proceedings of the {IEEE} International Conference on Big
  Data}, pages 4945--4954. IEEE, 2020.

\bibitem{Keogh2004ClusteringOT}
Eamonn~J. Keogh and Jessica Lin.
\newblock Clustering of time-series subsequences is meaningless: implications
  for previous and future research.
\newblock {\em Knowledge and Information Systems}, 8:154--177, 2004.

\bibitem{Ye2009TimeSS}
Lexiang Ye and Eamonn~J. Keogh.
\newblock Time series shapelets: a new primitive for data mining.
\newblock In {\em Proceedings of the {ACM SIGKDD} International Conference on
  Knowledge Discovery and Data Mining}, pages 947--956, 2009.

\bibitem{Zhu2016MatrixPI}
Yan Zhu, Zachary Schall-Zimmerman, Nader~Shakibay Senobari, Chin-Chia~Michael
  Yeh, Gareth~J. Funning, Abdullah~Al Mueen, Philip Brisk, and Eamonn~J. Keogh.
\newblock {Matrix Profile II:} exploiting a novel algorithm and {GPUs} to break
  the one hundred million barrier for time series motifs and joins.
\newblock {\em Proceedings of the {IEEE} International Conference on Data
  Mining}, pages 739--748, 2016.

\bibitem{wallace1968information}
Chris~S Wallace and David~M Boulton.
\newblock An information measure for classification.
\newblock {\em The Computer Journal}, 11(2):185--194, 1968.

\bibitem{seninge2021vega}
Lucas Seninge, Ioannis Anastopoulos, Hongxu Ding, and Joshua Stuart.
\newblock Vega is an interpretable generative model for inferring biological
  network activity in single-cell transcriptomics.
\newblock {\em Nature Communications}, 12(1):5684, 2021.

\bibitem{rissanen1978modeling}
Jorma Rissanen.
\newblock Modeling by shortest data description.
\newblock {\em Automatica}, 14(5):465--471, 1978.

\bibitem{Kingma2013AutoEncodingVB}
Diederik~P. Kingma and Max Welling.
\newblock Auto-encoding variational {Bayes}.
\newblock {\em CoRR}, abs/1312.6114, 2013.

\bibitem{JimenezRezende2014StochasticBA}
Danilo~Jimenez Rezende, Shakir Mohamed, and Daan Wierstra.
\newblock Stochastic backpropagation and approximate inference in deep
  generative models.
\newblock In {\em Proceedings of the International Conference on Machine
  Learning}, pages 1278--1286, 2014.

\bibitem{Maddison2016TheCD}
Chris~J. Maddison, Andriy Mnih, and Yee~Whye Teh.
\newblock The concrete distribution: A continuous relaxation of discrete random
  variables.
\newblock {\em ArXiv}, abs/1611.00712, 2016.

\bibitem{gumbel1954statistical}
Emil~Julius Gumbel.
\newblock {\em Statistical theory of extreme values and some practical
  applications: a series of lectures}, volume~33.
\newblock US Government Printing Office, 1954.

\bibitem{maddison2014sampling}
Chris~J Maddison, Daniel Tarlow, and Tom Minka.
\newblock A* sampling.
\newblock {\em Advances in Neural Information Processing systems}, 27, 2014.

\bibitem{shannon1959coding}
Claude~E Shannon et~al.
\newblock Coding theorems for a discrete source with a fidelity criterion.
\newblock {\em International Convention Record}, 4(142-163):1, 1959.

\bibitem{Foutsos1994FastSM}
Christos Faloutsos, M.~Ranganathan, and Yannis Manolopoulos.
\newblock Fast subsequence matching in time-series databases.
\newblock {\em {ACM SIGMOD} Record}, 23(2):419--429, 1994.

\bibitem{keogh2005exact}
Eamonn Keogh and Chotirat~Ann Ratanamahatana.
\newblock Exact indexing of dynamic time warping.
\newblock {\em Knowledge and Information Systems}, 7:358--386, 2005.

\bibitem{Lin2003ASR}
Jessica Lin, Eamonn~J. Keogh, Stefano Lonardi, and Bill~Yuan chi Chiu.
\newblock A symbolic representation of time series, with implications for
  streaming algorithms.
\newblock In {\em Proceedings of the Workshop on Research Issues on Data Mining
  and Knowledge Discovery}, 2003.

\bibitem{Popivanov2002SimilaritySO}
Ivan Popivanov and Ren{\'e}e~J. Miller.
\newblock Similarity search over time-series data using wavelets.
\newblock {\em Proceedings of the International Conference on Data
  Engineering}, pages 212--221, 2002.

\bibitem{song2020transitional}
Kiburm Song, Minho Ryu, and Kichun Lee.
\newblock Transitional sax representation for knowledge discovery for time
  series.
\newblock {\em Applied Sciences}, 10(19):6980, 2020.

\bibitem{chen2023efficient}
Xinye Chen and Stefan G{\"u}ttel.
\newblock An efficient aggregation method for the symbolic representation of
  temporal data.
\newblock {\em ACM Transactions on Knowledge Discovery from Data}, 17(1):1--22,
  2023.

\bibitem{rezvani2019new}
Roonak Rezvani, Payam Barnaghi, and Shirin Enshaeifar.
\newblock A new pattern representation method for time-series data.
\newblock {\em IEEE Transactions on Knowledge and Data Engineering},
  33(7):2818--2832, 2019.

\bibitem{li2022new}
Yucheng Li and Derong Shen.
\newblock A new symbolic representation method for time series.
\newblock {\em Information Sciences}, 609:276--303, 2022.

\bibitem{Yeh2016MatrixPI}
Chin-Chia~Michael Yeh, Yan Zhu, Liudmila Ulanova, Nurjahan Begum, Yifei Ding,
  Hoang~Anh Dau, Diego~Furtado Silva, Abdullah~Al Mueen, and Eamonn~J. Keogh.
\newblock {Matrix Profile I: All} pairs similarity joins for time series: A
  unifying view that includes motifs, discords and shapelets.
\newblock {\em Proceedings of the IEEE International Conference on Data
  Mining}, pages 1317--1322, 2016.

\bibitem{lu2022matrix}
Yue Lu, Renjie Wu, Abdullah Mueen, Maria~A Zuluaga, and Eamonn Keogh.
\newblock Matrix profile xxiv: scaling time series anomaly detection to
  trillions of datapoints and ultra-fast arriving data streams.
\newblock In {\em Proceedings of the 28th ACM SIGKDD Conference on Knowledge
  Discovery and Data Mining}, pages 1173--1182, 2022.

\bibitem{Alaee2021TimeSM}
Sara Alaee, Ryan Mercer, Kaveh Kamgar, and Eamonn~J. Keogh.
\newblock Time series motifs discovery under {DTW} allows more robust discovery
  of conserved structure.
\newblock {\em Data Mining and Knowledge Discovery}, 35:863--910, 2021.

\bibitem{imamura2023parameter}
Makoto Imamura and Takaaki Nakamura.
\newblock Parameter-free spikelet: Discovering different length and warped time
  series motifs using an adaptive time series representation.
\newblock In {\em Proceedings of the 29th ACM SIGKDD Conference on Knowledge
  Discovery and Data Mining}, pages 857--866, 2023.

\bibitem{noering2021pattern}
Fabian Kai-Dietrich Noering, Yannik Schroeder, Konstantin Jonas, and Frank
  Klawonn.
\newblock Pattern discovery in time series using autoencoder in comparison to
  nonlearning approaches.
\newblock {\em Integrated Computer-Aided Engineering}, 28(3):237--256, 2021.

\bibitem{Fortuin2018SOMVAEID}
Vincent Fortuin, Matthias H{\"u}ser, Francesco Locatello, Heiko Strathmann, and
  Gunnar R{\"a}tsch.
\newblock {SOM-VAE:} interpretable discrete representation learning on time
  series.
\newblock In {\em Proceedings of the International Conference on Learning
  Representations}, 2018.

\bibitem{higgins2017beta}
Irina Higgins, Loic Matthey, Arka Pal, Christopher Burgess, Xavier Glorot,
  Matthew Botvinick, Shakir Mohamed, and Alexander Lerchner.
\newblock {beta-VAE:} learning basic visual concepts with a constrained
  variational framework.
\newblock In {\em Proceedings of the International Conference on Learning
  Representations}, 2017.

\bibitem{Kirschbaum2018LeMoNADeLM}
Elke Kirschbaum, Manuel Haussmann, Steffen Wolf, Hannah Sonntag, Justus
  Schneider, Shehabeldin Elzoheiry, Oliver Kann, Daniel Durstewitz, and Fred~A.
  Hamprecht.
\newblock {LeMoNADe:} learned motif and neuronal assembly detection in calcium
  imaging videos.
\newblock {\em arXiv: Neurons and Cognition}, 2018.

\bibitem{makhzani2013k}
Alireza Makhzani and Brendan Frey.
\newblock K-sparse autoencoders.
\newblock {\em arXiv preprint arXiv:1312.5663}, 2013.

\bibitem{van2017neural}
Aaron Van Den~Oord, Oriol Vinyals, and Koray Kavukcuoglu.
\newblock Neural discrete representation learning.
\newblock {\em Advances in Neural Information Processing Systems}, 30, 2017.

\bibitem{becker2018interpreting}
S\"oren Becker, Marcel Ackermann, Sebastian Lapuschkin, Klaus-Robert M\"uller,
  and Wojciech Samek.
\newblock Interpreting and explaining deep neural networks for classification
  of audio signals.
\newblock {\em CoRR}, abs/1807.03418, 2018.

\bibitem{olszewski2001generalized}
Robert~Thomas Olszewski.
\newblock {\em Generalized feature extraction for structural pattern
  recognition in time-series data}.
\newblock Carnegie Mellon University, 2001.

\bibitem{thakoor2005hidden}
Ninad Thakoor, Sungyong Jung, and Jean Gao.
\newblock Hidden markov model based weighted likelihood discriminant for
  minimum error shape classification.
\newblock In {\em 2005 IEEE International Conference on Multimedia and Expo},
  pages 342--345. IEEE, 2005.

\bibitem{drews2008patient}
Frank~A Drews.
\newblock Patient monitors in critical care: Lessons for improvement.
\newblock {\em Advances in Patient Safety: New Directions and Alternative
  Approaches}, 3, 2008.

\bibitem{forde2014intentional}
Carol Forde-Johnston.
\newblock Intentional rounding: a review of the literature.
\newblock {\em Nursing Standard}, 28(32), 2014.

\bibitem{MatrixProfile2023}
Matrix~Profile Foundation.
\newblock What are time series snippets?, 2023.
\newblock Accessed: 2023-05-21.

\bibitem{kingma2014adam}
Diederik~P Kingma and Jimmy Ba.
\newblock Adam: A method for stochastic optimization.
\newblock {\em arXiv preprint arXiv:1412.6980}, 2014.

\bibitem{ismail2019deep}
Hassan Ismail~Fawaz, Germain Forestier, Jonathan Weber, Lhassane Idoumghar, and
  Pierre-Alain Muller.
\newblock Deep learning for time series classification: a review.
\newblock {\em Data mining and knowledge discovery}, 33(4):917--963, 2019.

\bibitem{hurley2009comparing}
Niall Hurley and Scott Rickard.
\newblock Comparing measures of sparsity.
\newblock {\em IEEE Transactions on Information Theory}, 55(10):4723--4741,
  2009.

\end{thebibliography}

\end{document}